%% file: icml2026_conference.tex
\documentclass{article}

\usepackage{microtype}
\usepackage{graphicx}
\usepackage{subcaption}
\usepackage{booktabs} 

\usepackage{hyperref}
\usepackage{xspace}

\newcommand{\method}{{\tt{PISCES}}\xspace}

\usepackage{algorithm}
\usepackage{algpseudocode}

\usepackage[accepted]{icml2026}

\usepackage{amsmath}
\usepackage{amssymb}
\usepackage{mathtools}
\usepackage{amsthm}

\usepackage{url}

\usepackage{amsfonts}       
\usepackage{nicefrac}       
\usepackage{bm}
\usepackage{multirow}
\usepackage{multicol}
\usepackage{makecell}
\usepackage[usestackEOL]{stackengine}
\usepackage{pifont}
\usepackage[dvipsnames]{xcolor}
\usepackage{wasysym}

\usepackage[capitalize,noabbrev]{cleveref}

\theoremstyle{plain}

\theoremstyle{definition}

\theoremstyle{remark}

\usepackage[textsize=tiny]{todonotes}
\makeatletter
\DeclareRobustCommand\onedot{\futurelet\@let@token\@onedot}
\def\@onedot{\ifx\@let@token.\else.\null\fi\xspace}
\def\eg{\emph{e.g}\onedot}

\makeatother

\icmltitlerunning{\method: Annotation-free Text-to-Video Post-Training via Optimal Transport-Aligned Rewards}

\begin{document}

\twocolumn[
  \icmltitle{\method: Annotation-free Text-to-Video Post-Training via\\Optimal Transport-Aligned Rewards}



  \icmlsetsymbol{equal}{*}
  \icmlsetsymbol{dagger}{$\dagger$}

  \begin{icmlauthorlist}
    \icmlauthor{Minh-Quan Le}{equal,comp,yyy}
    \icmlauthor{Gaurav Mittal}{equal,comp}
    \icmlauthor{Cheng Zhao}{comp}
    \icmlauthor{Xianfeng David Gu}{yyy}
    \icmlauthor{Dimitris Samaras}{yyy}
    \icmlauthor{Mei Chen}{dagger,comp}
  \end{icmlauthorlist}

  \icmlaffiliation{comp}{Microsoft}
  \icmlaffiliation{yyy}{Stony Brook University}


  \icmlkeywords{Machine Learning, ICML}
{
\includegraphics[width=\textwidth]{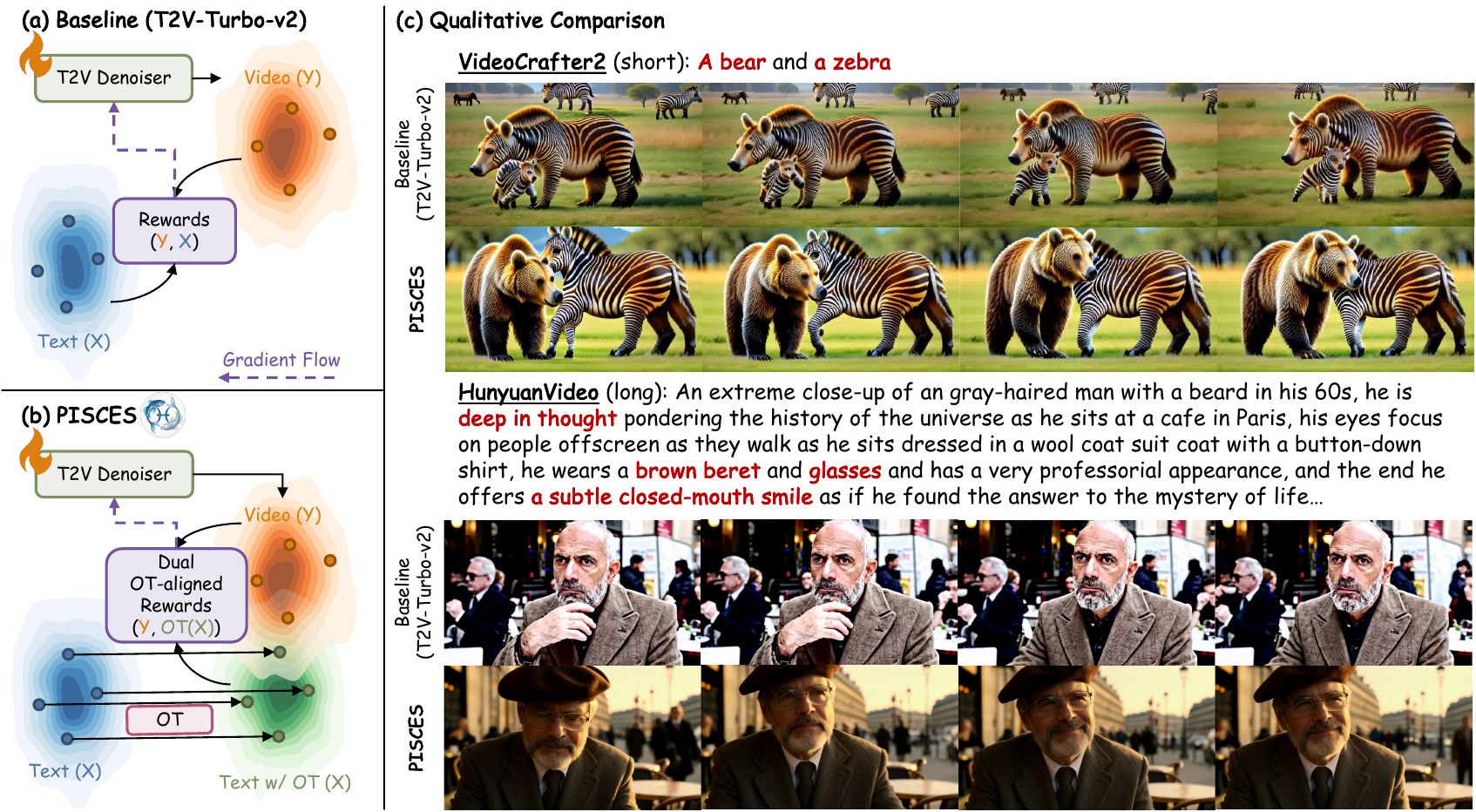}%
\vspace{-2mm}
\captionof{figure}{
\small(a) Baseline (T2V-Turbo-v2) defines rewards over pre-trained VLM text-video embeddings, which suffer from distributional misalignment. (b) \method T2V post-training addresses this by formulating reward supervision over an OT-aligned embedding space. We propose a novel Dual OT-aligned Rewards module that aligns text embeddings to the video space, enabling effective visual and semantic alignment. (c) Compared to the baseline, \method improves visual quality (temporal coherence, photorealism) and semantic fidelity (object count, attributes) on both short-video (VideoCrafter2) and long-video (HunyuanVideo) generation.
\label{fig:teaser}
\vspace{-5mm}
}
}
  \vskip 0.3in
]



\printAffiliationsAndNotice{\icmlEqualContribution}

\begin{abstract}
Text-to-video (T2V) generation aims to synthesize videos with high visual quality and temporal consistency that are semantically aligned with input text. Reward-based post-training has emerged as a promising direction to improve the quality and semantic alignment of generated videos. However, recent methods either rely on large-scale human preference annotations or operate on misaligned embeddings from pre-trained vision-language models, leading to limited scalability or suboptimal supervision. We present $\texttt{PISCES}$, an annotation-free post-training algorithm that addresses these limitations via a novel Dual Optimal Transport (OT)-aligned Rewards module. To align reward signals with human judgment, $\texttt{PISCES}$ uses OT to bridge text and video embeddings at both distributional and discrete token levels, enabling reward supervision to fulfill two objectives: (i) a Distributional OT-aligned Quality Reward that captures overall visual quality and temporal coherence; and (ii) a Discrete Token-level OT-aligned Semantic Reward that enforces semantic, spatio-temporal correspondence between text and video tokens. To our knowledge, $\texttt{PISCES}$ is the first to improve annotation-free reward supervision in generative post-training through the lens of OT. Experiments on both short- and long-video generation show that $\texttt{PISCES}$ outperforms both annotation-based and annotation-free methods on VBench across Quality and Semantic scores, with human preference studies further validating its effectiveness. We show that the Dual OT-aligned Rewards module is compatible with multiple optimization paradigms, including direct backpropagation and reinforcement learning fine-tuning.
Project page: \href{https://roar-ai.github.io/pisces}{https://roar-ai.github.io/pisces}
\end{abstract}
\input{sec/1_intro}
\input{sec/2_relatedwork}

\input{sec/3_preliminaries}
\input{sec/4_method}
\input{sec/5_experiments}

\input{sec/6_conclusion}

\section*{Acknowledgments}
This work was carried out during Minh-Quan’s internship at Microsoft. Dimitris Samaras was supported in part by NSF grants IIS-2123920 and IIS-2212046. Xianfeng David Gu was supported by NIH R21EB029733.
\section*{Impact Statement}

This work presents \method, a scalable and annotation-free post-training framework for improving text-to-video (T2V) generation via Dual OT-aligned Rewards. By reducing reliance on costly human annotations, \method offers a principled alternative to learning reward models from human preferences. Its distributional OT alignment strengthens global text–video coherence and visual quality, while its token-level OT alignment improves fine-grained semantic grounding, together enhancing both visual fidelity and semantic consistency. These improvements may broaden the usability of T2V models for beneficial applications such as education (visual explanations and instructional content), robotics and embodied AI (rapid prototyping of scenarios and simulations), and scientific visualization (communicating complex processes via controllable video synthesis).

More broadly, by making reward-based post-training more scalable, our approach enables researchers and practitioners to iterate on alignment methods without requiring large-scale preference datasets, potentially lowering barriers to experimentation and enabling wider participation. Because \method is annotation-free, it reduces the dependence on extensive human labeling pipelines, helping mitigate practical challenges related to cost, scalability, and the need to expose annotators to sensitive content.


\method is a training framework that improves alignment and quality using existing model signals, and it does not introduce any new categories of risk beyond those already associated with modern T2V systems. We emphasize that responsible deployment is important: applications should incorporate established safeguards (e.g., content moderation policies, provenance, or disclosure mechanisms when appropriate, and careful dataset and evaluation practices). We encourage future work to continue improving mechanisms for the safe deployment of T2V systems, particularly in high-impact or public-facing settings.


\bibliography{icml2026_conference}
\bibliographystyle{icml2026}

\input{sec/X_suppl}
\end{document}

%% file: sec/1_intro.tex
\vspace{-6mm}
\section{Introduction}
\label{sec:intro}
Text-to-video (T2V) generation~\citep{kong2025hunyuanvideosystematicframeworklarge, gen3} aims to synthesize videos from textual descriptions such that they appear realistic, temporally consistent, and accurately reflect the prompt. T2V has broad applications in multimedia content creation, robotics, and accessibility. While T2V performance is inherently subjective and judged by human preferences, recent benchmark~\citep{huang2024vbench} formalizes evaluation along two main dimensions -- \textit{Quality score}, for the visual quality and temporal consistency; and \textit{Semantic score}, factoring the correspondence of generated videos to text prompts.

Rapid advances in diffusion and flow matching models~\citep{podell2024sdxl, esser2024scaling} and Vision-Language Models (VLMs)~\citep{chung2023unimax, sun2024hunyuanlargeopensourcemoemodel, glm2024chatglmfamilylargelanguage} have enabled the development of recent T2V models~\citep{pika1,gen3,chen2024videocrafter2,kong2025hunyuanvideosystematicframeworklarge}. To further improve existing T2V models~\citep{chen2024videocrafter2, kong2025hunyuanvideosystematicframeworklarge}, particularly in terms of video-text misalignment in the denoisers, reward-based post-training~\citep{li2025tvturbov2, liu2025improving} has been introduced that provides additional supervision via specially designed rewards to the denoiser.

Reward-based T2V post-training methods can be either Annotation-based or Annotation-free. Annotation-based approaches~\citep{liu2025improving, yang2026dualipo, unifiedreward} collect large-scale human preference datasets, where annotators judge generated video pairs on quality and semantics, which are used to train a reward model or via Direct Preference Optimization (DPO)~\citep{rafailov2023direct, Wallace_2024_CVPR} for post-training. Although effective and serving as existing SoTA, these annotation-based methods cannot easily scale because they need high-quality preference-based annotations. Another line of work explores Annotation-free rewards, where supervision is derived from pre-trained VLMs rather than human labels~\citep{li2024t2vturbo, li2025tvturbov2}. While these approaches do not need large-scale human annotations, their performance is not on par with the Annotation-based techniques. We aim to achieve the best of both worlds by asking: \textit{Can an annotation-free T2V post-training method match, or even outperform, annotation-based approaches}?

From a review of annotation-free approaches, we identify reliance on pre-trained vision–language models (VLMs) for reward supervision as a key limitation. VLMs are trained with non-distributional objectives, such as pointwise matching \cite{chen2020uniter} and contrastive learning~\citep{radford2021clip}, that may not adequately align text with the real-video distribution, consistent with the patterns in Table~\ref{tab:OT_analysis} and Figure~\ref{fig:tsne}. This results in both quality and semantic issues, as shown in Figure~\ref{fig:teaser}c, such as failure to ensure the correct number and attributes of objects (\eg, “a zebra and a bear”, ``wearing a brown beret and glasses'') or failing to capture motion descriptors (\eg, ``closed-mouth smile'').

We posit that, for annotation-free reward supervision to mimic human preferences, the real-video distribution must be better aligned with the text distribution, which represents the space of human instructions/preferences, without altering the video distribution’s semantic structure, and the derived rewards should reflect human judgments of text-to-video outputs on the dual of quality and semantics. We introduce \method\footnote[1]{In astrology, Pisces is symbolized by two fish, signifying balance across realms. \method echoes this by aligning text and video through complementary quality and semantic rewards.}, an annotation-free T2V post-training algorithm that includes a novel\textbf{ Dual Optimal Transport-aligned Rewards} module~(Figure~\ref{fig:teaser}b). Leveraging Optimal Transport~(OT)~\citep{villani2009optimal, cuturi2013sinkhorn}, we tailor \method specifically for T2V post-training by enhancing text-video alignment at both the distribution and the token level to simultaneously improve both the visual quality and semantic consistency. For this, the Dual Rewards module comprises: (i) a \textbf{Distributional OT-aligned Quality Reward}, which learns a distributional OT map to transform text embeddings into the real-video embedding space while preserving their internal structure and enforcing temporal consistency and visual quality; and (ii) a \textbf{Discrete Token-level OT-aligned Semantic Reward}, which constructs a semantic spatio-temporal cost matrix over text and video tokens and solves a partial OT problem with an entropic Sinkhorn solver~\citep{cuturi2013sinkhorn}, to supervise correspondence by aligning text tokens with the most semantically, spatially, and temporally consistent video regions.

We validate \method on both short-video (VideoCrafter2~\citep{chen2024videocrafter2}) and long-video (HunyuanVideo~\citep{kong2025hunyuanvideosystematicframeworklarge}) generator~(Figure~\ref{fig:teaser}c) via VBench~\citep{huang2024vbench} as well as human evaluation. We show that our Dual OT-aligned Rewards module is applicable across different optimization paradigms, including direct backpropagation (gradient backpropagation through reward models) and reinforcement learning (RL) fine-tuning (GRPO~\citep{Guo2025DeepSeekR1, liu2025flowgrpo}). In doing so, we find that \method can significantly outperform all existing reward-based post-training approaches (both Annotation-based and Annotation-free). Through careful realignment of the text-video space, \method demonstrates that an annotation-free T2V post-training method can outperform Annotation-based approaches, making it a much stronger alternative at scale. Our key contributions are:

\begin{itemize}
    \item We introduce \method, a novel annotation-free post-training framework for T2V generation. For the first time, we identify a core bottleneck in existing VLM-based rewards operating on misaligned text-video embeddings and address this by leveraging OT to align embeddings, enabling reward supervision in a semantically meaningful, structure-preserving space.

    \item \method defines a novel Dual OT-aligned Rewards module comprising: (1) a \textit{Distributional OT-aligned Quality Reward}, capturing overall visual quality and temporal consistency; and (2) a \textit{Discrete Token-level OT-aligned Semantic Reward}, targeting localized text-video alignment for semantic consistency.
    
    \item \method outperforms both annotation-based and annotation-free T2V post-training methods on both Semantic and Quality Scores for short and long video generation, as validated by automatic metrics and human evaluations. We show that OT-aligned rewards are applicable to multiple optimization strategies.
\end{itemize}

%% file: sec/2_relatedwork.tex
\vspace{-1em}
\section{Related Work}
\label{sec:relatedwork}

\textbf{Reward-based Post-Training for T2V.}
In the image domain, reward models such as HPSv3~\citep{Ma_2025_ICCV}, ImageReward~\citep{xu2023image}, PickScore~\citep{kirstain2023pickapic}, and Hummingbird~\cite{le2025hummingbird} have proven effective for aligning generations with text prompts. Extending to video, annotation-based approaches train on large-scale human preference datasets, including VideoReward~\citep{liu2025improving}, Dual-IPO~\citep{yang2026dualipo}, and UnifiedReward~\citep{unifiedreward}. While effective, these methods incur high annotation costs and suffer from limited scalability. Orthogonally, annotation-free methods leverage verifiable rewards~\citep{le2025newtonrewards} or pre-trained VLMs such as ViCLIP~\citep{wang2023viclip} and InternVideo2~\citep{wang2024internvideo2}, with cosine-similarity rewards adopted in T2V-Turbo~\citep{li2024t2vturbo}, T2V-Turbo-v2~\citep{li2025tvturbov2}, while InstructVideo~\citep{Yuan_2024_CVPR} still relies on image-text rewards. However, these rewards operate on misaligned text–video embedding spaces, reducing their effectiveness to perform on par with annotation-based methods.
We identify this reward misalignment as the core bottleneck in T2V post-training and propose \method, the first to explore aligning embeddings via OT in generative post-training. Our framework introduces dual distributional and token-level OT-aligned rewards, enabling scalable and effective annotation-free supervision.

\textbf{Optimal Transport.}
Optimal Transport (OT) provides a principled framework for aligning probability distributions and has been widely applied in machine learning tasks such as domain adaptation~\citep{katageri2024synergizing}, generative modeling~\citep{tong2024improving, li2023dpm}, and cross-modal retrieval~\citep{han2024learning, izquierdo2024optimal}. Neural Optimal Transport (NOT)~\citep{korotin2023neural} further offers a scalable alternative by learning explicit transport maps via neural networks. Recent works have also leveraged discrete OT for alignment in vision tasks: \cite{xie2025discovering} formulate disentangled OT for visual--concept relations, while \cite{liu2025boosting} integrates OT into query reformation for temporal action localization. Other efforts include HOTS3D~\citep{li2024hots3d}, which aligns text and image features using spherical OT, and OT-CLIP~\citep{shi2024otclip}, which reframes CLIP training and inference as OT problems. Despite these advances, prior work has not explored OT in the context of reward modeling for generative T2V post-training. \method is the first to address misaligned pre-trained VLM text-video embeddings for annotation-free T2V rewards, and introduces a Dual OT-aligned Rewards module  capturing both distributional alignment and discrete token-level text-video correspondence.

%% file: sec/4_method.tex
\begin{figure*}[t!]
    \centering
    \vspace{-3mm}
    \includegraphics[width=0.95\linewidth]{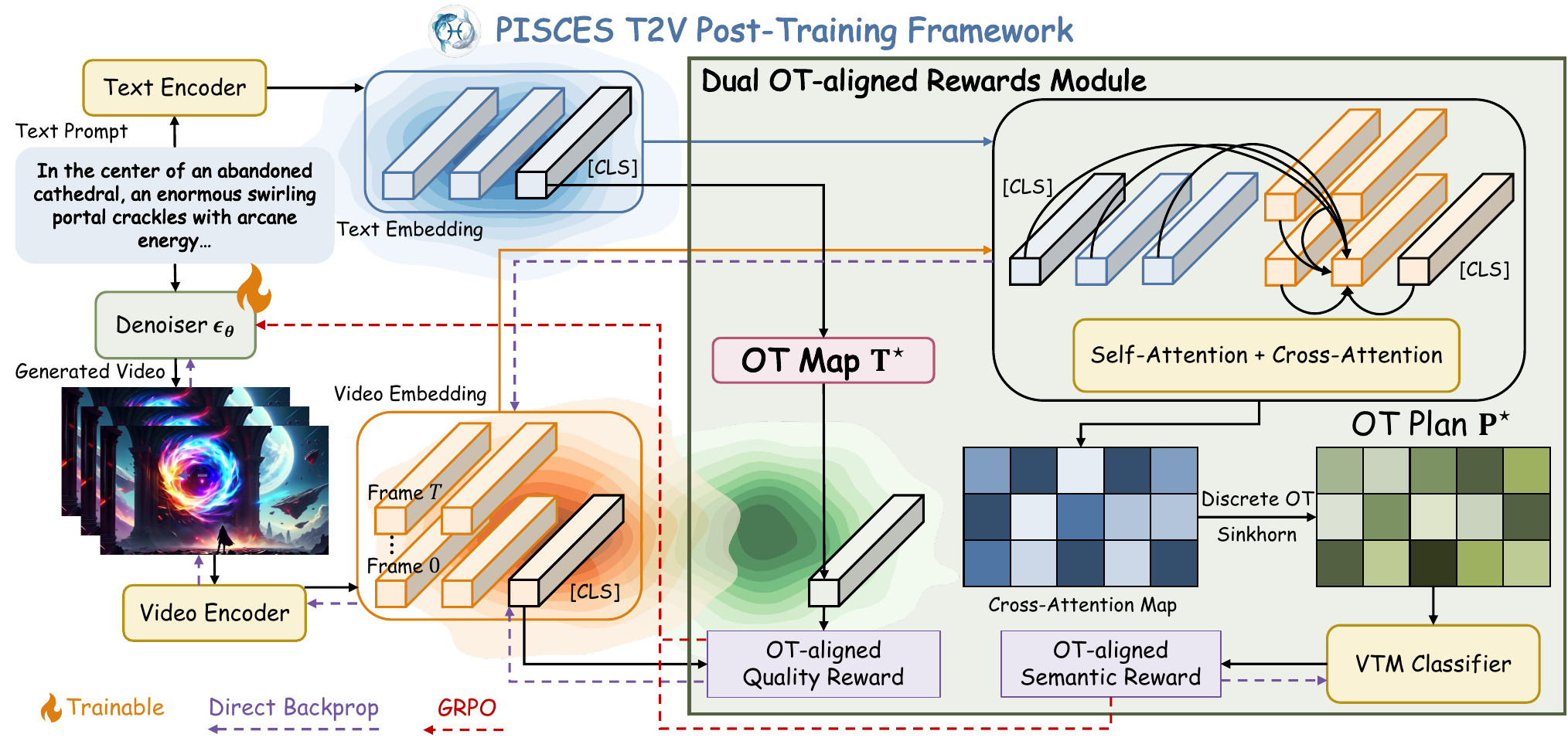}
\vspace{-2mm}
\caption{\method \textbf{T2V Post-Training.} We introduce a Dual OT-aligned Rewards module: (i) a distributional OT map $\mathbf{T}^\star$ for Quality Reward via {\tt{[CLS]}} representation similarity, and (ii) a discrete OT plan $\mathbf{P}^\star$ with spatio-temporal constraints for Semantic Reward via a Video-Text Matching (VTM) classifier. The rewards module provides supervision for fine-tuning the T2V denoiser and is applicable with direct backpropagation and RL fine-tuning (GRPO).}    
    \label{fig:overview}
    \vspace{-5mm}
\end{figure*}

\section{Method}
Fig~\ref{fig:overview} provides an overview of \method's Dual OT-aligned Rewards module and T2V post-training algorithm.

\subsection{Distributional OT-aligned Quality Reward}
When humans evaluate the \textit{quality} of a generated video, they attend to global properties such as realism, motion coherence, and overall visual consistency -- essentially asking whether the video could plausibly belong to the distribution of real-videos. To mimic this process in an annotation-free manner, we align text embeddings onto the manifold of real-video embeddings before defining the reward.

For this, we formulate the distributional alignment as a Monge–Kantorovich OT problem.  OT~\citep{villani2009optimal} provides a principled framework for aligning two probability distributions $\mu \in \mathcal{P}(\mathcal{Y})$ and $\nu \in \mathcal{P}(\mathcal{X})$ by finding a transport map $\mathbf{T}: \mathcal{Y} \rightarrow \mathcal{X}$ that pushes $\mu$ to $\nu$ (i.e., $\mathbf{T}_{\sharp}\mu = \nu$) while minimizing a transport cost $\mathbf{c}(\mathbf{y}, \mathbf{x})$. Given text embeddings $\mathcal{Y}$ and real-video embeddings $\mathcal{X}$ extracted from a pre-trained VLM, we train an OT map $\mathbf{T}: \mathcal{Y} \rightarrow \mathcal{X}$ using NOT~\citep{korotin2023neural} with the objective:
{\footnotesize
\begin{equation}
\sup_{f} \inf_{\mathbf{T}} \int_{\mathcal{X}} f(\mathbf{x}) d\nu(\mathbf{x}) + \int_{\mathcal{Y}} \left(\mathbf{c}(\mathbf{y}, \mathbf{T}(\mathbf{y})) - f(\mathbf{T}(\mathbf{y}))\right)d\mu(\mathbf{y}),
\end{equation}
}
where $\mathbf{c}(\mathbf{y}, \mathbf{x}) = \|\mathbf{y}-\mathbf{x}\|^2$ is the transport cost. We implement this via iterative optimization of the transport map $\mathbf{T}_\psi$ and potential function $f_\omega$ parameterized by neural networks, as shown in Algorithm~\ref{algorithm:OT-alignment} in Appendix~\ref{sec:OTalignment-appendix}. The resulting OT-aligned embeddings $\mathbf{T}^\star(\mathbf{y})$ reduce distributional mismatch while preserving the semantic structure of the embedding space (see Table~\ref{tab:OT_analysis} and Figure~\ref{fig:tsne}).

Once the text distribution is aligned with the real-video distribution, comparing an OT-aligned text embedding $\mathbf{T}^\star(\mathbf{y})$ with a generated video embedding $\hat{\mathbf{x}}$ becomes equivalent to comparing a real-video embedding $\mathbf{x}^{\mathrm{real}}$ with $\hat{\mathbf{x}}$. The OT map thus projects text embeddings into the real-video manifold, making $\mathbf{T}^\star(\mathbf{y})$ a proxy for $\mathbf{x}^{\mathrm{real}}$. With this intuition, we define the Quality Reward~(Fig~\ref{fig:overview}) as the cosine similarity between global representation {\tt [CLS]} tokens:
{\small
\begin{equation}
\mathcal{R}_\textrm{OT-quality} =
\frac{\mathbf{T}^\star(\mathbf{y}_{\mathrm{[CLS]}})^T \cdot \mathbf{\hat{x}}_{\mathrm{[CLS]}}}
{\|\mathbf{T}^\star(\mathbf{y}_{\mathrm{[CLS]}})\| \|\mathbf{\hat{x}}_{\mathrm{[CLS]}}\|}
\approx
\frac{(\mathbf{x}^{\mathrm{real}}_{\mathrm{[CLS]}})^T \cdot \mathbf{\hat{x}}_{\mathrm{[CLS]}}}
{\|\mathbf{x}^{\mathrm{real}}_{\mathrm{[CLS]}}\|\|\mathbf{\hat{x}}_{\mathrm{[CLS]}}\|}.
\end{equation}
}
Cosine similarity provides a natural coherence signal by comparing the \emph{direction} of embeddings, making it robust to scale or style differences while capturing structural consistency. After OT projects text embeddings into the real-video manifold, cosine becomes a geometry-respecting measure of quality, evaluating whether generated videos point in the same “quality direction” as real ones.

\subsection{Discrete Token-level OT-aligned Semantic Reward}
When judging semantic fidelity in T2V, humans implicitly ask whether the prompt's keywords are actually reflected in the generated video. To mimic this in T2V post-training, we introduce a token-level reward based on Partial OT (POT).

To facilitate a strong semantic alignment, we integrate discrete POT into the cross-attention layers of pre-trained VLM InternVideo2~\cite{wang2024internvideo2}. Vanilla cross-attention, however, often fails to capture precise multimodal correspondences: it operates directly on misaligned embeddings and distributes attention diffusely across irrelevant patches, as seen in Fig~\ref{fig:OT_plan}. Without a mechanism to enforce selective, structured grounding, important tokens may fail to connect to the right visual regions. To address this problem, we design a novel mechanism which augments attention with a POT-guided transport plan that enforces semantic, temporal, and spatial consistency between text and video tokens. For each cross-attention head, we construct a cost matrix between text tokens $\mathbf{y}$ and video patch tokens $\hat{\mathbf{x}}$ comprising three components specifically designed for T2V rewards:

\textbf{Semantic similarity:} $1 - \cos(\mathbf{y}_i, \hat{\mathbf{x}}_j)$, encouraging tokens with similar meaning to align.
    
\textbf{Temporal constraint:} $|\tau(\mathbf{y}_i) - t_j|$, where $\tau(\mathbf{y}_i) = \sum_k\mathbf{A}_{ik}*t_k$ is the expected frame index of text token $i$ under attention  $\mathbf{A}$, and $t_j$ is frame index of video patch $j$.
    
\textbf{Spatial constraint:} $|\pi(\mathbf{y}_i) - s_j|_2$, where $\pi(\mathbf{y}_i)=\sum_k\mathbf{A}_{ik} * s_k$ is the expected 2D position of text token $i$ (under attention $\mathbf{A}$) and $s_j$ is the spatial coordinate of video patch $j$ on the frame grid.

The final cost matrix is:
$\mathbf{C}_{ij} = \text{semantic}(i,j) + \gamma \cdot \text{temporal}(i,j) + \eta \cdot \text{spatial}(i,j)$,
with $\gamma,\eta$ balancing temporal and spatial penalties. We then solve a partial OT problem on this cost matrix $\mathbf{C}$ via an entropic Sinkhorn solver~\citep{cuturi2013sinkhorn} with fraction-of-mass $m=0.9$, as shown in Algorithm~\ref{alg:pot-sinkhorn}. This produces a transport plan $\mathbf{P}^\star$ that softly matches each text token to a subset of video tokens, rather than forcing full mass transport. To integrate this into InternVideo2, we propose to inject $\mathbf{P}^\star$ into the vanilla attention $\mathbf{A}$ via log-space fusion, a lightweight, differentiable mechanism.
This yields updated cross-attention probabilities $\tilde{\mathbf{A}}$ that combines standard attention with POT-guided structure. Formally, the updated cross-attention map is:
{\small
\begin{equation}
\tilde{\mathbf{A}} \propto \exp\!\big(\log(\mathbf{A} + \varepsilon) + \log(\mathbf{P}^\star + \varepsilon)\big).
\end{equation}
}
This fusion preserves differentiability through $\mathbf{A}$ while treating $\mathbf{P}^\star$ as a structural prior. Finally, the POT-refined features are passed into the pre-trained Video-Text Matching (VTM) classifier of InternVideo2, which outputs two logits for positive and negative matches. The positive logit after softmax ($\mathrm{idx} = 1$) provides the Semantic Reward:
{\small
\begin{equation}
\mathcal{R}_\textrm{OT-semantic} = \texttt{softmax}\left({\tt{VTM}}\left[\tilde{\mathbf{A}} \cdot \mathbf{\hat{x}}\right]\right)_{\mathrm{idx}=1}.
\end{equation}
}
Our discrete POT-based Semantic Reward captures human selectivity: not every word needs to be grounded, and important tokens are matched to relevant patches. The spatio-temporal cost further constrains \textit{where} and \textit{when} content should appear. Together, this provides a reward signal evaluating text–video correspondence in a human-aligned way. 


\subsection{Post-Training}

\noindent \textbf{Direct Backpropagation.} We integrate the Dual OT-aligned Rewards module into consistency distillation~\citep{song2023consistency, wang2023videolcm, lu2025simplifying} for efficient refinement, optimizing the denoiser:
\begin{equation}
\mathcal{L}_\textrm{direct} = \mathcal{L}_{\mathrm{CD}}(\boldsymbol{\theta}, \boldsymbol{\theta}^-;\phi) - \mathcal{R}_\textrm{OT-quality} - \mathcal{R}_\textrm{OT-semantic}.
\end{equation}
\noindent {\bf GRPO.} At each step, given a prompt $\mathbf{y}$, we sample a group of videos using SDEs $\{\mathbf{x}_0^1,\mathbf{x}_0^2,\dots, \mathbf{x}_0^G\}$ from the video denoiser $\pi_{\theta_{\textrm{old}}}$, and optimize the policy model $\pi_{\theta}$ with the objective~\citep{Guo2025DeepSeekR1, liu2025flowgrpo}: 
\begin{equation}
\small
\begin{aligned}
& \mathcal{L}_\textrm{GRPO} = \mathbb{E}_{\substack{\left\{\mathbf{x}_0^i\right\}_{i=1}^G \sim \pi_{\theta_{\text{old}}}(\cdot \mid \mathbf{y}) \\
\mathbf{a}_{t,i} \sim \pi_{\theta_{\text{old}}}\!\left(\cdot \mid \mathbf{s}_{t,i}\right)}}
\left[
\frac{1}{G} \sum_{i=1}^G \frac{1}{T} \sum_{t=1}^T
\right.\\
&\qquad\left.
\max\Bigl(-\rho_{t,i} A_i,\; -\operatorname{clip}\!\left(\rho_{t,i}, 1-\epsilon, 1+\epsilon\right) A_i\Bigr)
\right].
\end{aligned}
\end{equation}
where $\rho_{t, i}=\frac{\pi_\theta\left(\mathbf{a}_{t, i} \mid \mathbf{s}_{t, i}\right)}{\pi_{\theta_{o l d}}\left(\mathbf{a}_{t, i} \mid \mathbf{s}_{t, i}\right)}$ and $A_i = \frac{r_i - \textrm{mean}(\{r_1, r_2, \dots, r_G\})}{\textrm{std}(\{r_1, r_2, \dots, r_G\})}$ is the advantage function computed using a group of rewards $\{r_1, r_2, \dots, r_G\}$ with Dual OT-aligned Rewards module.

During post-training, we use LoRA~\citep{hu2022lora}, freezing all parameters except the denoiser. Algorithm~\ref{algorithm:fine-tuning} summarizes this procedure: we generate videos using Euler ODE for direct backprop and SDE for GRPO, compute the Dual OT-aligned Rewards, and optimize the denoiser.

\setlength{\textfloatsep}{8pt}
\setlength{\intextsep}{8pt}
\setlength{\floatsep}{6pt}
\begin{algorithm}[!t]
{\small
\caption{Post-Training with OT-aligned Rewards}
\label{algorithm:fine-tuning}
\begin{algorithmic}
\Require Pre-trained denoiser $\boldsymbol{\epsilon_\theta}$; data $\mathbf{p}_{\textrm{data}}$; ODE solver $\Phi$; skipping interval $k$; distance $d(\cdot,\cdot)$; $\boldsymbol{\theta}^- \leftarrow \boldsymbol{\theta}$
\Ensure $\boldsymbol{\epsilon_\theta}$ converges and minimizes $\mathcal{L}_\textrm{total}$
\While{not converged}
\State Sample video-text $(\mathbf{x}_{\textrm{video}},\mathbf{y}_{\textrm{text}}) \sim \mathbf{p}_{\textrm{data}}$, $n \sim \mathcal{U}[1, N-k]$
\State $\mathbf{z}_0 \leftarrow \mathcal{E}(\mathbf{x_{\textrm{video}}})$ \Comment{Encode $\mathbf{x}_{\textrm{video}}$ to latent space $\mathbf{z}_0$}
\State Extract text embedding $\mathbf{y}$ from $\mathbf{y}_{\textrm{text}}$
\State $\mathbf{z}_{t_{n +k}} \sim \mathcal{N}\left(\alpha(t_{n+k})\mathbf{z}_0; \beta^2(t_{n+k})\mathbf{I}\right)$ \Comment{Add noise to latent}
\State Perform ODE solver from $t_{n+k} \rightarrow t_n$: 
\State $\hat{\mathbf{z}}_{t_n}^\phi \leftarrow \mathbf{z}_{t_{n+k}} + (t_n - t_{n+k})\Phi(\mathbf{z}_{t_{n+k}}, t_{n+k};\phi)$ 
\State Compute Consistency Distillation loss:
\State $\mathcal{L}_{\mathrm{CD}}(\boldsymbol{\theta}, \boldsymbol{\theta}^-;\phi) \leftarrow d\left(\boldsymbol{g_\theta}(\mathbf{z}_{t_{n+k}}, t_{n+k}), \boldsymbol{g_{\theta^-}}(\hat{\mathbf{z}}_{t_n}^\phi, t_n)\right)$ 
\State Single-step ODE solver from $t_{n+k} \rightarrow 0$:
\State $\hat{\mathbf{z}}_{0}^\phi \leftarrow \mathbf{z}_{t_{n+k}} - \int_0^{t_{n+k}}\left(\gamma(t)\mathbf{z}_t + \frac{1}{2}\sigma^2(t)\boldsymbol{\epsilon_\theta}(\mathbf{z}_t, \mathbf{y}, t)\right)\mathrm{d}t$ \
\State $\hat{\mathbf{x}}_0 \leftarrow \mathcal{D}(\hat{\mathbf{z}}_{0}^\phi)$   \Comment{Decode $\hat{\mathbf{z}}_{0}^\phi$ to pixel space $\hat{\mathbf{x}}_0$}
\State Extract video embedding $\hat{\mathbf{x}}$ from video $\hat{\mathbf{x}}_0$ and compute rewards using $\hat{\mathbf{x}}$ and $\mathbf{y}$ with OT
\State $\mathcal{L}_\textrm{direct} = \mathcal{L}_{\mathrm{CD}}(\boldsymbol{\theta}, \boldsymbol{\theta}^-;\phi) - \mathcal{R}_\textrm{OT-quality} - \mathcal{R}_\textrm{OT-semantic}$ or $\mathcal{L}_\textrm{totalGRPO} = \mathcal{L}_{\mathrm{CD}}(\boldsymbol{\theta}, \boldsymbol{\theta}^-;\phi) + \mathcal{L}_\textrm{GRPO}$
\State Backward $\mathcal{L}_{\textrm{direct}}$ or $\mathcal{L}_{\textrm{totalGRPO}}$ to update $\boldsymbol{\theta}$ 
\State $\boldsymbol{\theta}^- \leftarrow \tt{stop\_grad}(  \lambda\boldsymbol{\theta} + (1 - \lambda)\boldsymbol{\theta^-})$
\EndWhile
\end{algorithmic}}
\end{algorithm}

%% file: sec/5_experiments.tex
\section{Experiments}
\label{sec:experiments-main}
\subsection{Experimental Setting}

\input{tables/finetuning_comparison}
\input{tables/t2v_comparison}

\textbf{Implementation Details.}
We validate \method by post-training VideoCrafter2 \citep{chen2024videocrafter2} (2s @ 8FPS) and HunyuanVideo \citep{kong2025hunyuanvideosystematicframeworklarge} (5s @ 25FPS), representing short- and long-video settings. We use base text-video embeddings from InternVideo2 \citep{wang2024internvideo2} for OT-based reward alignment. We perform post-training on $8\times$ A100~80GB GPUs for 2 days with a learning rate of $1e{-6}$, batch size $1$, and accumulation $32$ (direct backprop) or 4 days for GRPO (includes time for intermediate inference and visualization, actual training time is $\approx\!30$ hours for direct backpropagation and $\approx\!78$ hours for GRPO). 
Both the OT map $\mathbf{T}_{\psi}$ and critic $f_\omega$ are $3$-layer MLPs with ReLU activations and LayerNorm. To train Neural OT (NOT) map, we use video-text pairs with 8-frame clips, extracted using frozen InternVideo2~\citep{wang2024internvideo2}, and train on 1 A100 GPU for one day, equivalent to 24 A100 GPU-hours.

\textbf{Datasets.}
Following T2V-Turbo-v2~\citep{li2025tvturbov2}, we construct a balanced dataset mixing WebVid10M~\citep{bain2021frozen} and VidGen-1M~\citep{tan2024vidgen1mlargescaledatasettexttovideo}. We sample 2s clips at 8FPS for VideoCrafter2 (short-video) and 5s clips at 25FPS for HunyuanVideo (long-video). We resize frames to $512\times320$ for VideoCrafter2 and $848\times480$ for HunyuanVideo before training.

\textbf{Evaluation Metrics.} We evaluate \method with VBench \citep{huang2024vbench}, benchmarking T2V generation across 16 dimensions summarized into a \textit{Quality Score} (visual fidelity, temporal coherence, \eg, subject/background consistency, motion smoothness) and a \textit{Semantic Score} (fine-grained alignment to prompts, \eg, object presence, spatial relations, action correctness). The \textit{Total Score}, a weighted sum of the two, provides a holistic measure of video fidelity and semantic alignment. We also conduct user study on 400 prompts as per VideoReward~\citep{liu2025improving}.

\subsection{Comparison with Existing Methods}

\textbf{Automatic Evaluation on VBench.} 
Table~\ref{tab:compact-fine-tuning} compares \method with existing T2V post-training methods~\citep{li2024t2vturbo, li2025tvturbov2, liu2025improving, Liu_2025_CVPR, unifiedreward} on both short-video~(VideoCrafter2) and long-video~(HunyuanVideo) generation. We observe that \method significantly outperforms both annotation-based and annotation-free approaches, achieving the highest scores across all metrics -- Total, Quality, and Semantic. Table~\ref{tab:compact-t2v-vbench} provides a broader comparison with recent T2V models, both open-source~\citep{wang2023modelscope, kong2025hunyuanvideosystematicframeworklarge, chen2024videocrafter2, zhang2024show} and closed-source~\citep{pika1, klingai, gen3}. We can again observe that, when post-trained with \method, HunyuanVideo outperforms existing T2V models on automatic VBench evaluation. This highlights the benefit of our OT formulation to align the text-video embeddings in an off-the-shelf pre-trained VLM, which strengthens supervision for the T2V post-training Rewards module. This also highlights the effectiveness of our proposed OT-aligned Quality and Semantic Rewards in performing optimal T2V post-training. We also show that \method's OT-aligned Rewards module is applicable to different optimizations in Appendix~\ref{sec:different_optimization}.

\begin{figure}[t]
    \centering
    \includegraphics[width=\linewidth]{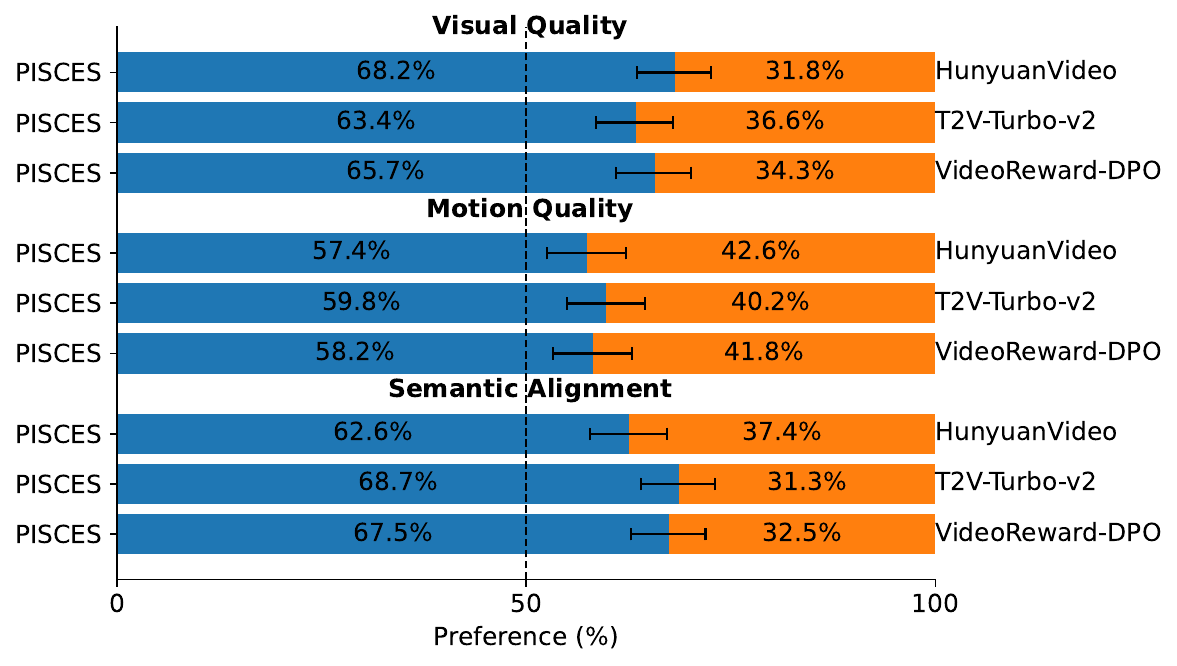}
    \vspace{-7mm}
    \caption{\small\textbf{Human preference study.}
\method outperforms HunyuanVideo, T2V-Turbo-v2, VideoReward-DPO in visual quality, motion and semantic alignment, validating its effectiveness in T2V.}
\label{fig:preference}
\vspace{-2mm}
\end{figure}


\textbf{Human Evaluation.} Following VideoReward~\citep{liu2025improving}, we conduct a human study on 400 prompts, evaluating three dimensions: visual quality, motion quality, and text alignment. For each prompt, we generate videos using pre-trained HunyuanVideo, post-trained T2V-Turbo-v2~\citep{li2025tvturbov2}, VideoReward-DPO~\citep{liu2025improving}, and \method. Participants ($n=85$) answer three questions per prompt: (1) Which video is better aligned with the text prompt? (2) Which video has better visual quality? and (3) Which video has better motion quality? As shown in Figure~\ref{fig:preference}, \method is consistently preferred over all baselines in terms of visual quality, motion quality, and semantic alignment. This confirms that our dual OT-aligned rewards enhance both visual and semantic consistency. Furthermore, experiments on evaluation prompts from VBench~\citep{huang2024vbench} and VideoReward~\citep{liu2025improving} imply the generalization of our approach to out-of-domain prompts, as we train T2V models only on WebVid10M~\citep{bain2021frozen} and VidGen-1M~\citep{tan2024vidgen1mlargescaledatasettexttovideo}.
\begin{figure*}[t]
    \centering
    \vspace{-3mm}
    \includegraphics[width=0.95\linewidth]{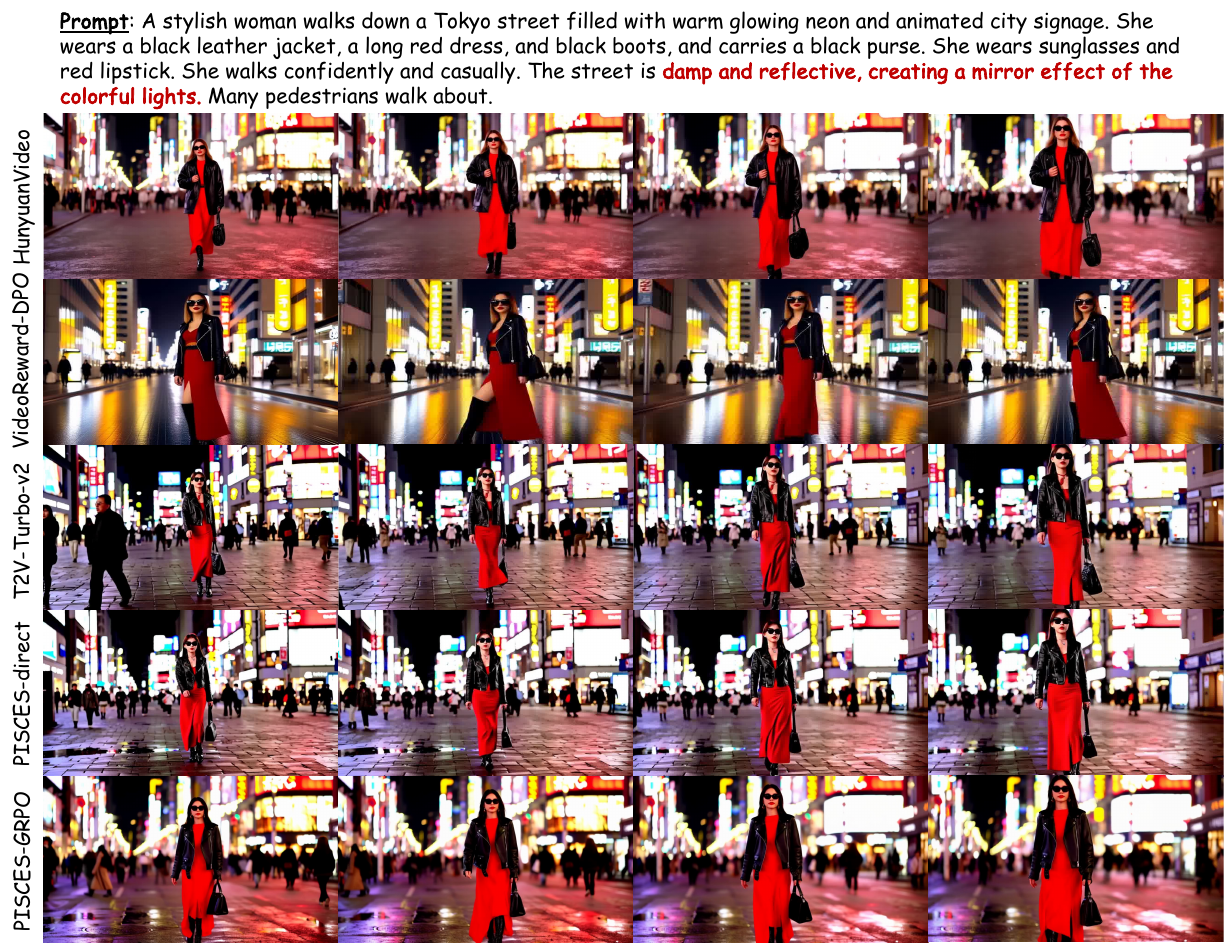}
    \vspace{-2mm}
    \caption{\small Qualitative comparison of T2V models. \method produces videos with better semantic fidelity and visual quality, accurately capturing key details such as the reflective wet pavement and vibrant neon lighting.}
    \label{fig:qualitative_comparison}
    \vspace{-5mm}
\end{figure*}

\textbf{Qualitative Comparison.} To further assess the effectiveness of \method, we visually compare videos generated by \method against other methods in Fig.~\ref{fig:qualitative_comparison}. Given the text prompt, we observe \method generates videos with improved semantic fidelity and visual coherence. Compared to baselines, \method is better at preserving fine-grained semantic details, such as the reflection effects on the wet pavement and the vibrant color contrast in the scene~(w.r.t. Semantic Score in VBench). The generated subject is also more globally consistent across frames, reducing temporal flicker and maintaining a stable appearance~(w.r.t. to Quality Score in VBench). These results align with our quantitative findings from the previous section, validating the effectiveness of Dual OT-aligned Rewards to enhance both visual quality and text-video alignment in T2V models.

\subsection{Ablation Study}

\input{tables/ablation_study}


\textbf{Effectiveness of OT in Text-Video Alignment.}
Table~\ref{tab:ablation_compact} (Rows 2 and 5) compares \method with and without OT on VideoCrafter2, isolating the effect of aligning text and video embedding distributions via OT. For the Quality Reward, we replace the OT-aligned $\mathbf{T}^\star(\mathbf{y}_{[\text{CLS}]})$ in Eq.~(2) with the raw $\mathbf{y}_{[\text{CLS}]}$, and for the Semantic Reward, we remove the POT–guided attention $\tilde{\mathbf{A}}$ in Eq.~(4) and use vanilla cross-attention $\mathbf{A}$. We find that OT is critical for improving both Quality and Semantic performance. \method attains a Semantic Score of $77.63$, outperforming the variant without OT ($75.82$), demonstrating that distribution and token-level alignment prior to reward computation substantially enhances semantic correspondence. OT also improves Quality Score ($83.73$ vs. $83.44$), confirming its role in structuring the feature space for more effective T2V post-training. Overall, these results show that OT-aligned embeddings provide stronger supervision than off-the-shelf pre-trained VLM embeddings used in existing post-training methods.


\begin{figure}[t]
    \centering
    \vspace{-2mm}
    \includegraphics[width=\linewidth]{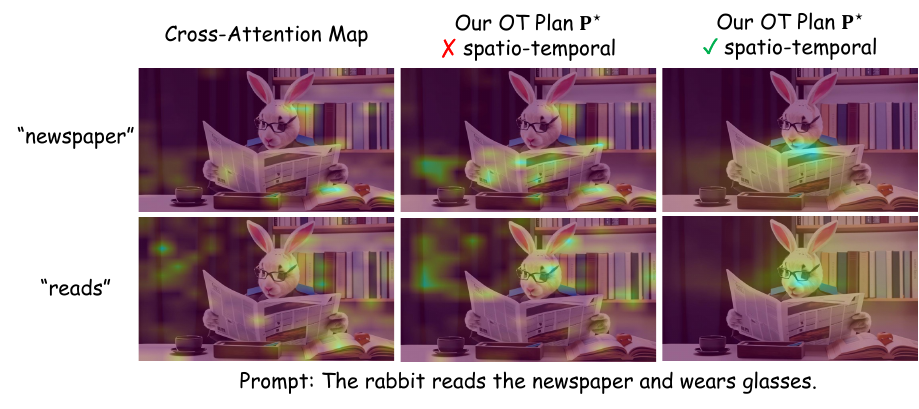}
    \vspace{-7mm}
    \caption{\small  Cross-attention maps (left) are diffuse, OT plan without spatio-temporal constraints (middle) misaligns tokens, while our constrained OT plan (right) produces accurate correspondences.}
    \label{fig:OT_plan}
    \vspace{-2mm}
\end{figure}

\textbf{Impact of OT-aligned Quality and Semantic Rewards.} Table~\ref{tab:ablation_compact} analyzes the contribution of each OT-aligned reward. Using only the Quality Reward (Row 3) improves the VBench Quality Score from $82.20$ to $83.77$ (Row 1), demonstrating its effectiveness in enhancing global coherence and visual quality. In contrast, using only the Semantic Reward (Row 4) substantially increases the Semantic Score from $73.42$ to $76.99$, highlighting its ability to capture fine-grained text–video alignment such as object presence and actions. Combining both rewards (Row 5) yields the best overall performance, improving both Quality and Semantic Scores over single-reward variants. This confirms their complementary roles within \method’s OT-aligned space. For results across all 16 VBench dimensions, see Appendix~\ref{sec:full-ablation}.

\subsection{Optimal Transport Analysis}
To validate the effectiveness of OT~\citep{villani2009optimal, cuturi2013sinkhorn} in aligning text and video embeddings, we conduct both qualitative and quantitative analyses. We extract $10,000$ text-video pairs from WebVid10M~\citep{bain2021frozen} using pre-trained VLM InternVideo2.


\textbf{OT Plan.} Figure~\ref{fig:OT_plan} compares standard cross-attention maps with our token-level OT plans. While cross-attention alone produces diffuse activations, and unconstrained OT plans misalign tokens, incorporating spatio-temporal constraints in the OT cost matrix yields meaningful correspondences. This highlights the benefit of discrete OT in our Semantic Reward: it ensures fine-grained alignment of text tokens with semantically and spatio-temporally consistent video regions, directly improving localized supervision during post-training. We further validate that our designed discrete POT helps improve the video-text matching performance of pre-trained InternVideo2 by $8.11\%$ in Appendix~\ref{sec:pot_analysis}.  

\input{tables/alignment_methods}


\textbf{Quantitative Analysis.} Mutual KNN~\citep{huh2024position} measures cross-modal alignment by evaluating k-nearest-neighbor overlap between text and video embeddings, where higher values indicate stronger alignment. Spearman correlation $r$ assesses structural preservation by measuring rank consistency before and after alignment. As shown in Table~\ref{tab:OT_analysis}, OT achieves the highest Mutual KNN and Spearman correlation among all methods, indicating effective distribution alignment with minimal structural distortion. Post-training HunyuanVideo with OT-aligned rewards further yields the best Quality and Semantic Scores, confirming improvements in both visual fidelity and text–video consistency.

\begin{figure}[t]
    \centering
    \vspace{-2mm}
    \includegraphics[width=\linewidth]{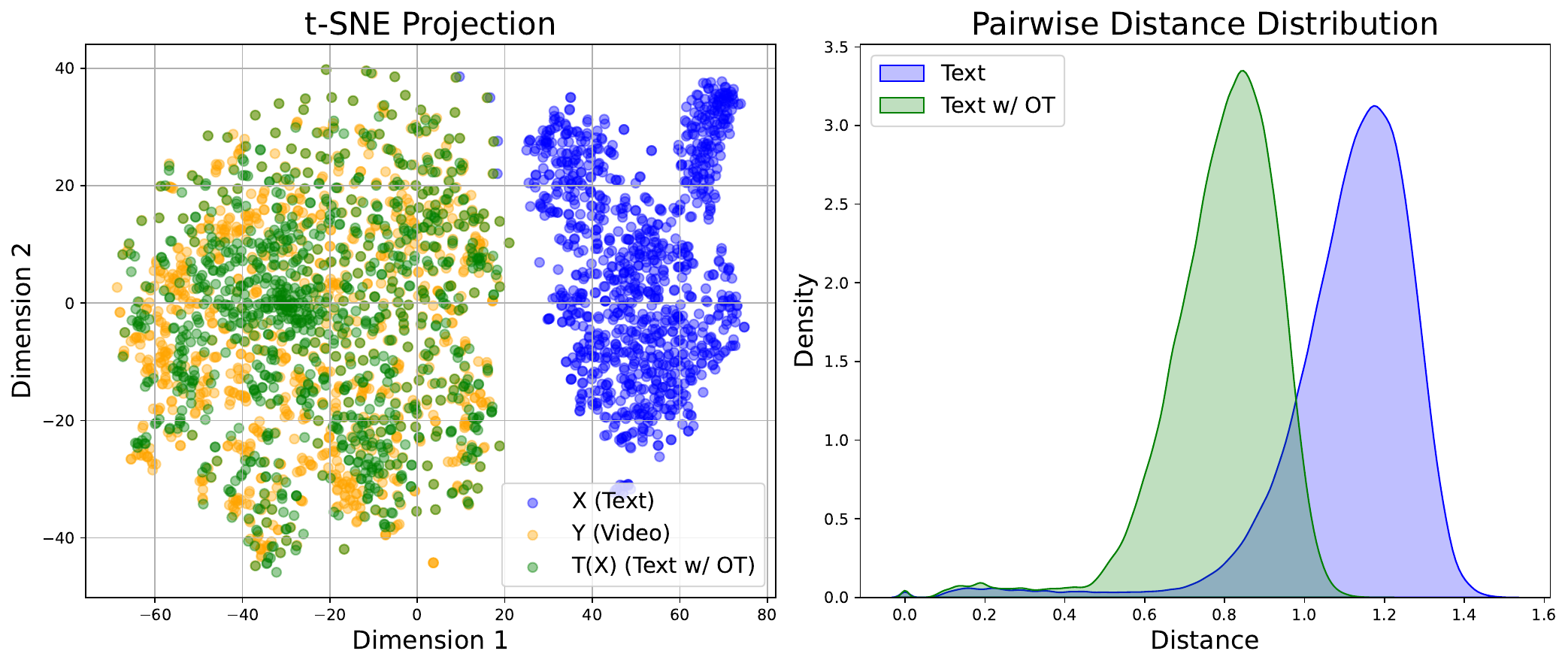}
    \vspace{-6mm}
    \caption{\small t-SNE shows OT aligns text embeddings (\textcolor{ForestGreen}{green}) closer to video embeddings distribution (\textcolor{orange}{orange}). Pairwise distance distribution indicates OT preserves text embedding structure.}
    \label{fig:tsne}
    \vspace{-2mm}
\end{figure}

\textbf{OT Map.} Fig.~\ref{fig:tsne} (left) presents a t-SNE projection of text and video embeddings. The original text embeddings (\textcolor{blue}{blue}) are largely misaligned with video embeddings (\textcolor{orange}{orange}), highlighting distributional gaps in VLMs (leading to suboptimal text-video alignment). OT-transformed text embeddings (\textcolor{ForestGreen}{green}) shift significantly closer to video embeddings, demonstrating improved alignment. Fig.~\ref{fig:tsne} (right) further supports this observation via pairwise distance distribution. The distribution of text embeddings after OT transformation closely resembles the original one, confirming OT aligns embeddings without distorting their internal relationships.

\begin{figure}[t]
    \centering
    \includegraphics[width=\linewidth]{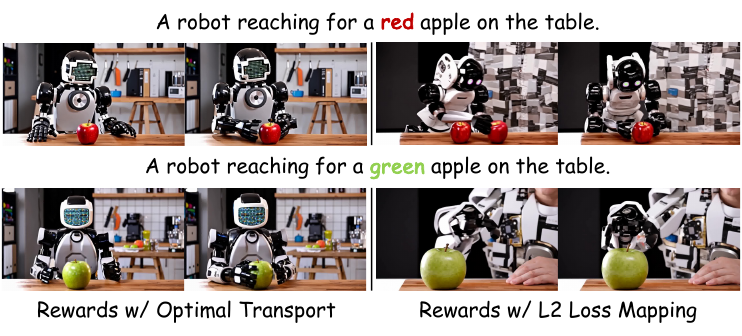}
    \vspace{-6mm}
    \caption{\small  Post-training with OT-aligned rewards (left) yields consistent outputs with expected color changes, while L2 loss mapping (right) causes sampling inconsistencies and visual artifacts.}
    \vspace{-2mm}
    \label{fig:OT_linear}
\end{figure}
\textbf{Impact on T2V Post-Training.} In Figure~\ref{fig:OT_linear}, we conduct a controlled experiment using the prompt \textit{``A robot reaching for a \textcolor{red}{red}/\textcolor{ForestGreen}{green} apple on the table''} with the same random seed. Post-training with OT-aligned rewards (left) maintains structural consistency -- background, robot appearance, and stable motion, with only the apple color changing as expected. In contrast, L2 loss mapping (right) produces unstable outputs: robot appearance and object placement vary unpredictably, and artifacts such as disappearing objects emerge. These results confirm that distorted text embeddings distribution harms reward-based post-training, leading to unstable sampling and degraded video quality. Overall, these findings underscore the advantages of OT alignment. By ensuring rewards operate on a well-structured space, OT prevents distributional distortions, resulting in improved sampling stability and higher-quality T2V generation.

%% file: tables/finetuning_comparison.tex
\begin{table*}[!t]
\vspace{-1mm}
\caption{\small \textbf{VBench comparison on VideoCrafter2 and HunyuanVideo}. \method significantly outperforms existing reward-based T2V post-training methods across all scores. *Reproduced without motion guidance for fair comparison. Additional analysis in Appendix~\ref{sec:motion_guide}.}
\vspace{-2mm}
\centering
\setlength\tabcolsep{4pt}

\resizebox{0.8\linewidth}{!}{
\begin{tabular}{@{}l|c@{~~}c@{~~}c|c@{~~}c@{~~}c@{}}
\toprule
\multirow{2}{*}{\textbf{Models}} & \multicolumn{3}{c|}{\textbf{VideoCrafter2}~\citep{chen2024videocrafter2}} & \multicolumn{3}{c}{\textbf{HunyuanVideo}~\citep{kong2025hunyuanvideosystematicframeworklarge}} \\
 & \textbf{Total} & \textbf{Quality} & \textbf{Semantic} & \textbf{Total} & \textbf{Quality} & \textbf{Semantic} \\
\midrule
Vanilla &80.44\xspace\xspace\xspace\xspace\xspace\xspace\xspace\xspace&82.20\xspace\xspace\xspace\xspace\xspace\xspace\xspace\xspace&73.42\xspace\xspace\xspace\xspace\xspace\xspace\xspace\xspace&83.24\xspace\xspace\xspace\xspace\xspace\xspace\xspace\xspace&85.09\xspace\xspace\xspace\xspace\xspace\xspace\xspace\xspace&75.82\xspace\xspace\xspace\xspace\xspace\xspace\xspace\xspace\xspace\\
VCM~\citep{wang2023videolcm} & 73.97 {\scriptsize \color{red}$\downarrow$6.47} & 78.54 {\scriptsize \color{red}$\downarrow$3.66} & 55.66 {\scriptsize \color{red}$\downarrow$17.8} & 81.77 {\scriptsize \color{red}$\downarrow$1.47} & 84.60 {\scriptsize \color{red}$\downarrow$0.49} & 70.49 {\scriptsize \color{red}$\downarrow$5.33} \\
T2V-Turbo~\citep{li2024t2vturbo} & 81.01 {\scriptsize \color{ForestGreen}$\uparrow$0.57} & 82.57 {\scriptsize \color{ForestGreen}$\uparrow$0.37} & 74.76 {\scriptsize \color{ForestGreen}$\uparrow$1.34} & 83.86 {\scriptsize \color{ForestGreen}$\uparrow$0.62} & 85.57 {\scriptsize \color{ForestGreen}$\uparrow$0.48} & 77.00 {\scriptsize \color{ForestGreen}$\uparrow$1.18} \\
T2V-Turbo-v2*~\citep{li2025tvturbov2} 
& 81.87 {\scriptsize \color{ForestGreen}$\uparrow$1.43} 
& 83.26 {\scriptsize \color{ForestGreen}$\uparrow$1.06} 
& 76.30 {\scriptsize \color{ForestGreen}$\uparrow$2.88} 
& 84.25 {\scriptsize \color{ForestGreen}$\uparrow$1.01} 
& 85.93 {\scriptsize \color{ForestGreen}$\uparrow$0.84} 
& 77.52 {\scriptsize \color{ForestGreen}$\uparrow$1.70} \\


VideoReward-DPO~\citep{liu2025improving} 
& 80.75 {\scriptsize \color{ForestGreen}$\uparrow$0.31} 
& 82.11 {\scriptsize \color{red}$\downarrow$0.09} 
& 75.29 {\scriptsize \color{ForestGreen}$\uparrow$1.87} 
& 83.54 {\scriptsize \color{ForestGreen}$\uparrow$0.30} 
& 85.02 {\scriptsize \color{red}$\downarrow$0.07} 
& 77.63 {\scriptsize \color{ForestGreen}$\uparrow$1.81} \\
VideoDPO~\citep{Liu_2025_CVPR} 
& 81.93 {\scriptsize \color{ForestGreen}$\uparrow$1.49} 
& 83.07 {\scriptsize \color{ForestGreen}$\uparrow$0.87} 
& 77.38 {\scriptsize \color{ForestGreen}$\uparrow$3.96} 
& 84.13 {\scriptsize \color{ForestGreen}$\uparrow$0.89} 
& 85.71 {\scriptsize \color{ForestGreen}$\uparrow$0.62} 
& 77.83 {\scriptsize \color{ForestGreen}$\uparrow$2.01} \\
UnifiedReward~\citep{unifiedreward} 
& 81.43 {\scriptsize \color{ForestGreen}$\uparrow$0.99}
& 83.26 {\scriptsize \color{ForestGreen}$\uparrow$1.06}
& 74.12 {\scriptsize \color{ForestGreen}$\uparrow$0.70}
& 83.80 {\scriptsize \color{ForestGreen}$\uparrow$0.56}
& 85.46 {\scriptsize \color{ForestGreen}$\uparrow$0.37}
& 77.15 {\scriptsize \color{ForestGreen}$\uparrow$1.33}\\

\method
& \textbf{82.75} {\scriptsize \color{ForestGreen}$\uparrow$\textbf{2.31}} 
& \textbf{84.05} {\scriptsize \color{ForestGreen}$\uparrow$\textbf{1.85}} 
& \textbf{77.54} {\scriptsize \color{ForestGreen}$\uparrow$\textbf{4.12}} 
& \textbf{85.45} {\scriptsize \color{ForestGreen}$\uparrow$\textbf{2.21}} 
& \textbf{86.73} {\scriptsize \color{ForestGreen}$\uparrow$\textbf{1.64}} 
& \textbf{80.33} {\scriptsize \color{ForestGreen}$\uparrow$\textbf{4.51}} \\
\bottomrule
\end{tabular}
}
\label{tab:compact-fine-tuning}
\vspace{-3mm}
\end{table*}

%% file: tables/t2v_comparison.tex
\begin{table*}[!t]
\caption{\small\textbf{Automatic Evaluation on VBench}. We compare different T2V models across Quality, Semantic, and Total Scores. HunyuanVideo, post-trained with \method, performs the best on all scores across all models.}
\vspace{-1mm}
\centering

\resizebox{\linewidth}{!}{
\begin{tabular}{l|ccccccc|cc}
\toprule
\multirow{2}{*}{\textbf{Metric}}  &  \multirow{2}{*}{\makecell{\textbf{ModelScope}\\ \citep{wang2023modelscope}}} &  \multirow{2}{*}{\makecell{\textbf{Show-1}\\ \citep{zhang2024show}}} & \multirow{2}{*}{\makecell{\textbf{Pika-1.0}\\ \citep{pika1}}} & \multirow{2}{*}{\makecell{\textbf{Gen-3}\\ \citep{gen3}}} & \multirow{2}{*}{\makecell{\textbf{Kling}\\ \citep{klingai}}} & \multirow{2}{*}{\makecell{\textbf{VideoCrafter2}\\ \citep{chen2024videocrafter2}}} & \multirow{2}{*}{\makecell{\textbf{HunyuanVideo}\\ \citep{kong2025hunyuanvideosystematicframeworklarge}}} & \multicolumn{2}{c}{\method} \\
 & & & & & & & & \textbf{VideoCrafter2} & \textbf{HunyuanVideo} \\
\midrule
\textbf{Quality Score} & 78.05 & 80.42 & 82.92 & 84.11 & 83.39 & 82.20 & 85.09 & 83.73 & \textbf{86.73} \\
\textbf{Semantic Score} & 66.54 & 72.98 & 71.77 & 75.17 & 75.68 &  73.42 & 75.82 & 77.63  & \textbf{80.33} \\
\textbf{Total Score} & 75.75 & 78.93 & 80.69 & 82.32 & 81.85 &  80.44 & 83.24 & 82.51 & \textbf{85.45} \\
\bottomrule
\end{tabular}
}
\label{tab:compact-t2v-vbench}
\vspace{-6mm}
\end{table*}

%% file: tables/ablation_study.tex
\begin{table}[!t]
\caption{\small\textbf{Ablation Study.} OT alignment improves both Quality and Semantic Scores, while Quality Reward enhances visual quality and Semantic Reward improves text-video correspondence. Full \method achieves the \textbf{best} performance.}
\vspace{-2mm}
\centering
\resizebox{\linewidth}{!}{
\begin{tabular}{l|c|c|c|c||c|c}
\toprule
\textbf{Method} & \textbf{OT} & $\mathcal{R}_{\textrm{semantic}}$ & $\mathcal{R}_{\textrm{quality}}$ & \makecell{\textbf{Total}\\\textbf{Score}} & \makecell{\textbf{Quality}\\\textbf{Score}} & \makecell{\textbf{Semantic}\\\textbf{Score}} \\
\midrule
Vanilla (VideoCrafter2) & \ding{55} & \ding{55} & \ding{55} & 80.44 & 82.20 & 73.42\\ 
\method w/o OT & \ding{55} & \ding{51} & \ding{51} & 81.92 & 83.44 & 75.82 \\ 
\method + $\mathcal{R}_{\textrm{OT-quality}}$  & \ding{51} & \ding{55} & \ding{51} & \underline{82.21} & \textbf{83.77} & 75.97 \\ 
\method + $\mathcal{R}_{\textrm{OT-semantic}}$ & \ding{51} & \ding{51} & \ding{55} & 81.70 & 82.87 & \underline{76.99} \\ 
\method Full & \ding{51} & \ding{51} & \ding{51} & \textbf{82.51} & \underline{83.73} & \textbf{77.63} \\
\bottomrule
\end{tabular}
}
\label{tab:ablation_compact}
\vspace{-3mm}
\end{table}

%% file: tables/alignment_methods.tex
\begin{table}[!t]
\caption{\small Comparison of alignment methods. OT improves text-video alignment (higher Mutual KNN) and preserves text embedding structure (higher Spearman Correlation). Post-training with OT-aligned rewards achieves the best Quality and Semantic Scores.}
\vspace{-2mm}
\centering
\resizebox{\linewidth}{!}{
\begin{tabular}{l|c|c|c|c}
\toprule
\textbf{Method} & \textbf{\makecell{Mutual\\KNN $\uparrow$}} & \textbf{\makecell{Spearman\\Correlation $\uparrow$}} & \textbf{\makecell{Quality\\Score $\uparrow$}} & \textbf{\makecell{Semantic\\Score $\uparrow$}} \\
\midrule
Contrastive & 0.2135 & - & \underline{85.57} & 77.00 \\
Mapping w/ L2 & \underline{0.2318} & \underline{0.4873} & 84.89 & 77.12 \\
Mapping w/ KL & 0.2284 & 0.4720 & 84.72 & \underline{77.15} \\
OT (\method) & \textbf{0.2597} & \textbf{0.9018} & \textbf{86.73} & \textbf{80.33} \\
\bottomrule
\end{tabular}
}
\vspace{-2mm}
\label{tab:OT_analysis}
\end{table}

%% file: sec/6_conclusion.tex
\section{Conclusion}
We introduce \method, the first annotation-free post-training T2V algorithm that outperforms all existing annotation-based and annotation-free methods on VBench and human evaluation. Overcoming the limitation of existing annotation-free methods, which rely on VLM embeddings misaligned at both distributional and token levels, \method introduces a novel Dual OT-aligned Rewards module through the lens of OT. It comprises a Distributional OT-aligned Quality Reward and a Discrete token-level Semantic Reward to significantly improve visual quality and semantic consistency across short- and long-video generation. \method paves the way for scalable, principled post-training in T2V and offers a general blueprint for OT-based reward design in multimodal generation.

%% file: sec/X_suppl.tex
\newpage
\appendix
\section*{Appendix}
In this technical appendix, we provide additional details, ablations, and analyses that support the main findings of our work. Section~\ref{sec:CD} introduces the Consistency Distillation (CD) \cite{song2023consistency, wang2023videolcm, lu2025simplifying} mechanism used to efficiently integrate OT-aligned rewards into the \method post-training process. Section~\ref{sec:OTalignment-appendix} describes the Neural Optimal Transport (NOT) \cite{villani2009optimal, korotin2023neural} formulation for aligning the distributions of text and video embeddings, while Section~\ref{sec:discrete-ot-appendix} presents our discrete token-level OT optimization via the entropic unbalanced Sinkhorn algorithm \cite{cuturi2013sinkhorn}.

Section~\ref{sec:different_optimization} shows that our dual reward module is compatible with different optimization paradigms (direct backpropagation and GRPO \cite{Guo2025DeepSeekR1, liu2025flowgrpo}). Section~\ref{sec:full-ablation} provides a detailed ablation study on the impact of OT alignment and reward types across the full 16 dimensions of VBench \cite{huang2024vbench}. Section~\ref{sec:pot_analysis} quantifies the impact of partial OT and structured spatio-temporal constraints on matching accuracy. Section~\ref{sec:motion_guide} studies the role of motion guidance during post-training.

Section~\ref{sec:viclip_evaluator} evaluates \method under ViCLIP-based \cite{wang2023viclip} reward models to confirm generalizability across video-text encoders. Section~\ref{sec:ood_prompts} reports alignment performance on out-of-distribution prompts. Section~\ref{sec:hyper_sensitivity} presents a hyperparameter sweep over the OT loss weights $(\gamma,\eta)$ to assess sensitivity. Section~\ref{sec:failure_mode} highlights a failure case of partial OT due to base encoder limitations. Section~\ref{sec:human_details} reports inter-rater reliability and checks for category-level bias in human evaluation. Section~\ref{sec:reward_fusion} discusses an adaptive weighting strategy for reward fusion. Section~\ref{sec:reward_interaction} assesses the interaction of the dual OT-aligned rewards during training. Section~\ref{sec:reward_hacking} analyzes reward hacking risks and shows how CD loss mitigates them. Finally, Section~\ref{sec:training_cost} reports GPU-hour cost and runtime efficiency compared to baseline methods.

\section{Consistency Distillation}
\label{sec:CD}
Consistency Models (CMs)~\citep{song2023consistency,wang2023videolcm, lu2025simplifying} improve efficiency by enforcing self-consistency in the PF-ODE trajectory. A learned function $\boldsymbol{g}: (\mathbf{z}_t, t) \mapsto \mathbf{z}_\epsilon$ satisfies $\boldsymbol{g}(\mathbf{z}_t, t) = \boldsymbol{g}(\mathbf{z}_{t'}, t'), \quad \forall t, t' \in [\epsilon, T]$.
A pre-trained diffusion model is distilled into a CM via the Consistency Distillation loss: 
{\small
\begin{equation}
\mathcal{L}_{\mathrm{CD}}(\boldsymbol{\theta}, \boldsymbol{\theta}^-; \phi) = \mathbb{E}_{\mathbf{z}, t} \left[d\left(\boldsymbol{g_\theta}(\mathbf{z}_{t+k}, t+k), \boldsymbol{g}_{\theta^-}(\hat{\mathbf{z}}_{t_n}^\phi, t_n)\right)\right],\end{equation}}
where an ODE solver $\Phi$ estimates $\hat{\mathbf{z}}_{t_n}^\phi \leftarrow \mathbf{z}_{t_{n+k}} + (t_n - t_{n+k})\Phi(\mathbf{z}_{t_{n+k}}, t_{n+k}; \phi)$. To stabilize learning, EMA updates $\boldsymbol{\theta}^- \leftarrow \tt{stop\_grad}(\lambda \boldsymbol{\theta} + (1 - \lambda) \boldsymbol{\theta}^-)$.
Consistency distillation allows reward fine-tuning to backpropagate via single-step denoising, approximating multi-step denoising to enhance both efficiency and supervision signal.

\section{Distributional Alignment with OT}
\label{sec:OTalignment-appendix}
We first address the embeddings distribution misalignment in pre-trained VLMs by formulating the alignment as an OT problem. Specifically, given text embeddings $\mathcal{Y}$ and real video embeddings $\mathcal{X}$ extracted from a pre-trained VLM, we learn an OT map $\mathbf{T}: \mathcal{Y} \rightarrow \mathcal{X}$ using NOT~\citep{korotin2023neural}. This OT mapping transforms text embeddings into a semantically-aligned space with video embeddings, significantly reducing the inherent distributional mismatch (see Figure~\ref{fig:overview} left). We define the OT problem as:
\begin{equation}
\sup_{f} \inf_{\mathbf{T}} \int_{\mathcal{X}} f(\mathbf{x}) d\nu(\mathbf{x}) + \int_{\mathcal{Y}} \left(\mathbf{c}(\mathbf{y}, \mathbf{T}(\mathbf{y})) - f(\mathbf{T}(\mathbf{y}))\right)d\mu(\mathbf{y}),
\end{equation}
where the cost function is the squared Euclidean distance, $\mathbf{c}(\mathbf{y}, \mathbf{x}) = \|\mathbf{y}-\mathbf{x}\|^2$. We implement this via iterative optimization of the transport map $\mathbf{T}_\psi$ and potential function $f_\omega$ parameterized by neural networks, as shown in Algorithm~\ref{algorithm:OT-alignment}. The resulting OT-aligned embeddings $\mathbf{T}^\star(\mathbf{y})$ preserve original embedding structure while aligning distributions.
\begin{algorithm}[!h]
\small
\caption{Text-Video Embeddings Distribution Alignment w/ OT}
\label{algorithm:OT-alignment}
\begin{algorithmic}
\Require  text and video embedding distributions $\boldsymbol{\mu}, \boldsymbol{\nu}$; mapping network $\mathbf{T}_\psi: \mathcal{Y} \rightarrow \mathcal{X}$; \\potential network $f_\omega: \mathcal{X} \rightarrow \mathbb{R}$; number of inner iterations $K_T$; cost function $\mathbf{c}: \mathcal{Y}\times\mathcal{X}\rightarrow\mathbb{R}$
\Ensure learned stochastic OT map $\mathbf{T}_\psi$ representing an OT plan between distributions $\boldsymbol{\mu}, \boldsymbol{\nu}$
\While{not converged}
\State {\tt{unfreeze}}($\mathbf{T}_\psi$); {\tt{freeze}}($f_\omega$) \Comment{$\mathbf{T}$ optimization}
\For{$k_T = 1, 2, \dots, K_T$}
    \State Sample a batch of text embeddings $Y \sim \boldsymbol{\mu}$
    \State $\mathcal{L}_\mathbf{T} \leftarrow \frac{1}{\vert Y \vert}\sum_{\mathbf{y} \in Y}\left[\mathbf{c}(\mathbf{y}, \mathbf{T}_\psi(\mathbf{y})) - f_\omega(\mathbf{T}_\psi(\mathbf{y}))\right]$
    \State Backward $\mathcal{L}_\mathbf{T}$ and update $\psi$ using $\frac{\partial\mathcal{L}_\mathbf{T}}{\partial\psi}$
\EndFor
\State {\tt{freeze}}($\mathbf{T}_\psi$); {\tt{unfreeze}}($f_\omega$) \Comment{$f$ optimization}
\State Sample batch of video and text embeddings $X \sim \boldsymbol{\nu}, Y \sim \boldsymbol{\mu}$ 
\State $\mathcal{L}_f\leftarrow\frac{1}{\vert Y\vert}\sum_{\mathbf{y}\in\mathcal{Y}}f_\omega(\mathbf{T}_\psi(\mathbf{y})) - \frac{1}{\vert X\vert}\sum_{\mathbf{x}\in X}f_\omega(\mathbf{x})$
\State Backward $\mathcal{L}_f$ and update $\omega$ using $\frac{\partial\mathcal{L}_f}{\partial\omega}$
\EndWhile

\end{algorithmic}
\end{algorithm}

\section{Discrete OT for Semantic Alignment}
\label{sec:discrete-ot-appendix}

\noindent\textbf{Problem setup.}
Given text tokens $\{\mathbf{y}_i\}_{i=1}^N$ and video patch tokens $\{\mathbf{x}_j\}_{j=1}^M$ from a cross-attention layer, let $\mathbf{A}\in\mathbb{R}_+^{N\times M}$ denote the vanilla attention (row-normalized), $t_j$ the frame index of patch $j$, and $s_j\in\mathbb{R}^2$ its spatial coordinate on a $h\times w$ grid. We construct a spatio-temporal, semantics-aware cost matrix $\mathbf{C}\in\mathbb{R}_+^{N\times M}$ as
\[
\small
\mathbf{C}_{ij} \;=\; \underbrace{1-\cos\!\big(\mathbf{y}_i,\mathbf{x}_j\big)}_{\text{semantic}}
\;+\; \gamma\,\underbrace{\big|\tau(\mathbf{y}_i)-t_j\big|}_{\text{temporal}}
\;+\; \eta\,\underbrace{\big\|\pi(\mathbf{y}_i)-s_j\big\|_2}_{\text{spatial}},
\]
where $\tau(\mathbf{y}_i)=\sum_{k} A_{ik}\,t_k$ and $\pi(\mathbf{y}_i)=\sum_{k} A_{ik}\,s_k$ are attention-weighted expectations of frame index and spatial position, respectively. Each component is range-normalized to $[0,1]$, followed by a min–max normalization of $\mathbf{C}$ to $[0,1]$ for numerical stability.

\medskip
\noindent\textbf{Partial entropic OT.}
Let uniform marginals $\boldsymbol{\mu}\!=\!\frac{1}{N}\mathbf{1}_N$ and $\boldsymbol{\nu}\!=\!\frac{1}{M}\mathbf{1}_M$. We solve a \emph{partial} OT problem via an entropic, unbalanced Sinkhorn objective:
\[
\begin{aligned}
\min_{\mathbf{P}\ge 0}\;&\;\langle \mathbf{P},\mathbf{C}\rangle
+\epsilon\sum_{i,j}\mathbf{P}_{ij}(\log \mathbf{P}_{ij}-1) \\
\text{s.t.}\;&\;\mathbf{P}\mathbf{1}_M \approx \tau_a\,\boldsymbol{\mu},\qquad
\mathbf{P}^\top\mathbf{1}_N \approx \tau_b\,\boldsymbol{\nu}.
\end{aligned}
\]
where $\epsilon>0$ is the entropic temperature and $(\tau_a,\tau_b)\in(0,1]$ relax the marginals to achieve an effective transported fraction $m\in(0,1]$ (we use $m\!=\!0.9$). This yields a soft plan $\mathbf{P}^\star$ that \emph{selectively} matches informative text tokens to consistent video regions, avoiding over-forced alignments. Algorithm~\ref{alg:pot-sinkhorn} describes our detailed implementation of solving partial OT using entropic (unbalanced) Sinkhorn.

\medskip
\noindent\textbf{Attention fusion (structure prior).}
We fuse $\mathbf{P}^\star$ with vanilla attention $\mathbf{A}$ in log-space:
\[
\tilde{\mathbf{A}} \;\propto\; \exp\!\Big(\log(\mathbf{A}+\varepsilon)\;+\;\log(\mathbf{P}^\star+\varepsilon)\Big),
\]
with small $\varepsilon>0$ for stability. Gradients flow through $\mathbf{A}$ while $\mathbf{P}^\star$ acts as a detached structural prior.

\medskip
\noindent\textbf{Semantic reward.}
Let ${\tt VTM}$ be the pre-trained video–text matching head. Using $\tilde{\mathbf{A}}$ to aggregate patch features, the Semantic Alignment Reward is
\[
\mathcal{R}_{\mathrm{OT\text{-}semantic}}
\;=\;
\mathrm{softmax}\Big(\,{\tt VTM}\big[\tilde{\mathbf{A}}\cdot\mathbf{\hat{x}}\big]\Big)_{(\mathrm{idx}=1)}.
\]
In practice, POT is applied per head and per cross-attention layer. We set $(\gamma,\eta)\!=\!(0.2,0.2)$, $\epsilon\!=\!0.05$, and $m\!=\!0.9$. Please refer to Appendix~\ref{sec:pot_analysis} for hyperparameter selection and ablation study.

\begin{algorithm}[!h]
\small
\caption{Partial OT via Entropic (Unbalanced) Sinkhorn}
\label{alg:pot-sinkhorn}
\begin{algorithmic}
\Require Cost matrix $\mathbf{C}\in\mathbb{R}_+^{N\times M}$ (normalized to $[0,1]$), entropic temperature $\epsilon>0$, target transported fraction $m\in(0,1]$, max iterations $K$, tolerance $\delta$
\Ensure Transport plan $\mathbf{P}^\star\in\mathbb{R}_+^{N\times M}$
\State \textbf{Uniform marginals:} $\boldsymbol{\mu}\!=\!\frac{1}{N}\mathbf{1}_N,\; \boldsymbol{\nu}\!=\!\frac{1}{M}\mathbf{1}_M$
\State \textbf{Map partial mass to unbalanced strength:} 
\Statex \hspace{2mm} $\rho \leftarrow 
\begin{cases}
\infty & \text{if } m \ge 0.999 \\
\epsilon \cdot \frac{m}{1-m} & \text{otherwise}
\end{cases}$,
\Statex \hspace{2mm}$\tau(\rho) \leftarrow 
\begin{cases}
1 & \text{if } \rho=\infty \\
\frac{\rho}{\rho+\epsilon} & \text{else}
\end{cases}$
\State \textbf{Set relaxations:} $\tau_a \leftarrow \tau(\rho),\; \tau_b \leftarrow \tau(\rho)$
\State \textbf{Log-kernel:} $\log\mathbf{K} \leftarrow -\mathbf{C}/\epsilon$
\State Initialize $\log\mathbf{u}\leftarrow \mathbf{0}_N,\; \log\mathbf{v}\leftarrow \mathbf{0}_M$;\quad $\log\boldsymbol{\mu}\leftarrow \log(\boldsymbol{\mu})$, $\log\boldsymbol{\nu}\leftarrow \log(\boldsymbol{\nu})$
\For{$k=1$ to $K$}
  \State $\log(\mathbf{K}\mathbf{v}) \leftarrow \log\sum\nolimits_j \exp\!\big(\log\mathbf{K}_{:,j} + \log\mathbf{v}_j\big)$ 
  \State $\log\mathbf{u}_{\text{new}} \leftarrow \tau_a \cdot \big(\log\boldsymbol{\mu} - \log(\mathbf{K}\mathbf{v})\big)$
  \State $\log(\mathbf{K}^\top\mathbf{u}_{\text{new}}) \leftarrow \log\sum\nolimits_i \exp\!\big(\log\mathbf{K}_{i,:} + \log\mathbf{u}_{\text{new},i}\big)$
  \State $\log\mathbf{v}_{\text{new}} \leftarrow \tau_b \cdot \big(\log\boldsymbol{\nu} - \log(\mathbf{K}^\top\mathbf{u}_{\text{new}})\big)$
  \If{$\max\{ \|\log\mathbf{u}_{\text{new}}-\log\mathbf{u}\|_\infty,\;\|\log\mathbf{v}_{\text{new}}-\log\mathbf{v}\|_\infty \} < \delta$}
     \State \textbf{break}
  \EndIf
  \State $\log\mathbf{u}\leftarrow\log\mathbf{u}_{\text{new}},\quad \log\mathbf{v}\leftarrow\log\mathbf{v}_{\text{new}}$
\EndFor
\State \textbf{Recover plan:} $\log\mathbf{P} \leftarrow \log\mathbf{u}\mathbf{1}_M^\top \;+\; \log\mathbf{K} \;+\; \mathbf{1}_N \log\mathbf{v}^\top$
\State $\mathbf{P}^\star \leftarrow \exp(\log\mathbf{P})$
\end{algorithmic}
\end{algorithm}

\section{Automatic Evaluation with Different Optimization Paradigms}
\label{sec:different_optimization}
We optimize the same OT-aligned reward under two training routes: (i) direct backpropagation through the reward models and (ii) RL fine-tuning via GRPO. As reported in Table~\ref{tab:different_optimization}, both procedures yield consistent gains over the vanilla baselines on VBench (Total/Quality/Semantic) for VideoCrafter2 and HunyuanVideo, and the resulting scores are comparable across paradigms. This indicates that our reward provides meaningful supervision signals whose benefits are largely agnostic to the optimization routine.

\begin{table}[!t]
\caption{\small \textbf{Automatic VBench comparison on VideoCrafter2 and HunyuanVideo}. \method post-training with direct backpropagation or RL fine-tuning GRPO achieves comparable performance and shows strong improvement over the pre-trained models.}
\centering
\setlength\tabcolsep{4pt}

\resizebox{\linewidth}{!}{
\begin{tabular}{@{}l|c@{~~}c@{~~}c|c@{~~}c@{~~}c@{}}
\toprule
\multirow{2}{*}{\textbf{Models}} & \multicolumn{3}{c|}{\textbf{VideoCrafter2}~\citep{chen2024videocrafter2}} & \multicolumn{3}{c}{\textbf{HunyuanVideo}~\citep{kong2025hunyuanvideosystematicframeworklarge}} \\
 & \textbf{Total} & \textbf{Quality} & \textbf{Semantic} & \textbf{Total} & \textbf{Quality} & \textbf{Semantic} \\
\midrule
Vanilla &80.44\xspace\xspace\xspace\xspace\xspace\xspace\xspace\xspace&82.20\xspace\xspace\xspace\xspace\xspace\xspace\xspace\xspace&73.42\xspace\xspace\xspace\xspace\xspace\xspace\xspace\xspace&83.24\xspace\xspace\xspace\xspace\xspace\xspace\xspace\xspace&85.09\xspace\xspace\xspace\xspace\xspace\xspace\xspace\xspace&75.82\xspace\xspace\xspace\xspace\xspace\xspace\xspace\xspace\xspace\\
\method (Direct Backpropagation) & 82.51 {\scriptsize \color{ForestGreen}$\uparrow$2.07} & 83.73 {\scriptsize \color{ForestGreen}$\uparrow$1.53} & \textbf{77.63} {\scriptsize \color{ForestGreen}$\uparrow$\textbf{4.21}} & 85.05 {\scriptsize \color{ForestGreen}$\uparrow$1.81} & \textbf{86.84} {\scriptsize \color{ForestGreen}$\uparrow$\textbf{1.75}} & 77.89 {\scriptsize \color{ForestGreen}$\uparrow$2.07} \\
\method (GRPO)
& \textbf{82.75} {\scriptsize \color{ForestGreen}$\uparrow$\textbf{2.31}} 
& \textbf{84.05} {\scriptsize \color{ForestGreen}$\uparrow$\textbf{1.85}} 
& 77.54 {\scriptsize \color{ForestGreen}$\uparrow$4.12} 
& \textbf{85.45} {\scriptsize \color{ForestGreen}$\uparrow$\textbf{2.21}} 
& 86.73 {\scriptsize \color{ForestGreen}$\uparrow$1.64} 
& \textbf{80.33} {\scriptsize \color{ForestGreen}$\uparrow$\textbf{4.51}} \\
\bottomrule
\end{tabular}
}
\label{tab:different_optimization}
\end{table}

\section{Comprehensive Ablation Study}
\label{sec:full-ablation}
\textbf{Effectiveness of OT Alignment.}
Comparing \method w/o OT (which post-trains with pre-trained VLM embeddings) against full \method in Table~\ref{tab:ablation-full}, we observe significant improvements in both Quality Score ($83.44 \rightarrow 83.73$) and Semantic Score ($75.82 \rightarrow 77.63$). This confirms that aligning text-video distributions before post-training enhances both global coherence and fine-grained text-video correspondence.

\textbf{Impact of Quality and Semantic Alignment Rewards.}
To assess the individual effects of OT-aligned Quality and Semantic Reward, we compare \method w/ $\mathcal{R}_{\textrm{OT-quality}}$ and $\mathcal{R}_{\textrm{OT-semantic}}$ separately. Quality Reward primarily improves global coherence, reflected in gains in Quality Score (83.77), Aesthetic Quality (66.92), and Subject Consistency (97.07). Meanwhile, Semantic Reward enhances fine-grained alignment, leading to improvements in Semantic Score (76.99), Human Action (96.60), and Spatial Relation (44.97).

\textbf{Full \method.}
Integrating both rewards (full \method) results in the highest Total Score (82.51). Notably, Overall Consistency (29.10) and Temporal Style (26.97) also improve, reinforcing that the combination of OT alignment and both rewards provides the best optimization signal for text-video post-training.

\begin{table*}[!t]
\centering
\caption{\small Ablation Study. We analyze the contributions of OT alignment and OT-aligned Rewards to post-training performance. \method w/o OT post-trains with pre-trained VLM embeddings, while \method w/ $\mathcal{R}_{\textrm{OT-semantic}}$ and \method with $\mathcal{R}_{\textrm{OT-semantic}}$ assess the impact of OT-aligned Quality and Semantic Rewards, respectively. Full \method, which integrates OT alignment and both rewards, achieves the best performance across Total, Quality, and Semantic Scores, demonstrating the effectiveness of structured reward optimization. Bold numbers denote the best results in each category.}
\setlength\tabcolsep{3pt}
\begin{center}

\resizebox{\linewidth}{!}{
\begin{tabular}{l|c|c|c|c||c|c|c|c|c|c|c|c}
\toprule
\textbf{Method} & \textbf{OT} & $\mathcal{R}_{\textrm{semantic}}$ & $\mathcal{R}_{\textrm{quality}}$ & \textbf{\Centerstack{Total\\Score}} & \textbf{\Centerstack{Quality\\Score}} & {\Centerstack{Subject\\Consist.}} & {\Centerstack{BG \\Consist.}} & 
{\Centerstack{Temporal\\Flicker.}} & {\Centerstack{Motion\\Smooth.}} & {\Centerstack{Aesthetic\\Quality}} & {\Centerstack{Dynamic\\Degree}} & {\Centerstack{Image\\Quality}} \\
\midrule
VideoCrafter2 & \ding{55} & \ding{55} & \ding{55} & 80.44 & 82.20 & 96.85 & 98.22 & 98.41 & 97.73 & 63.13 & 42.50  & 67.22 \\ 
\method w/o OT & \ding{55} & \ding{51} & \ding{51} & 81.92 & 83.44 & 96.99 & 97.66 & 98.02 & 97.16 & 66.39 & 52.78  & 70.50 \\ 
\method w/ $\mathcal{R}_{\textrm{OT-quality}}$  & \ding{51} & \ding{55} & \ding{51} & 82.21 & \textbf{83.77} & 97.07 & 97.58 & 97.72 & 97.10 & \textbf{66.92} & \textbf{58.06} & \textbf{70.56} \\ 
\method w/ $\mathcal{R}_{\textrm{OT-semantic}}$ & \ding{51} & \ding{51} & \ding{55} & 81.70 & 82.87 & \textbf{97.59} & \textbf{98.61} & 98.06 & \textbf{97.31} & 66.83 & 40.00  & 70.19 \\ 
\method & \ding{51} & \ding{51} & \ding{51} &\textbf{82.51} & 83.73 & 96.61 & 97.49 & \textbf{98.72} & 96.80 & 66.07 & \textbf{57.50} & 70.39 \\
\bottomrule
\end{tabular}
}
\resizebox{\linewidth}{!}{
\begin{tabular}{l|c|c|c|c|c|c|c|c|c|c|c|c|c}
\toprule
\textbf{Method} & \textbf{OT} & $\mathcal{R}_{\textrm{semantic}}$ & $\mathcal{R}_{\textrm{quality}}$ &  \textbf{\Centerstack{Semantic\\Score}}  &{\Centerstack{Object\\Class}}  & {\Centerstack{Multiple\\Objects}} & {\Centerstack{Human\\Action}} & {Color} & {\Centerstack{Spatial\\Relation.}} & {Scene} & {\Centerstack{Appear.\\Style}} & {\Centerstack{Temporal\\Style}} & {\Centerstack{Overall\\Consist.}} \\
\midrule
VideoCrafter2 & \ding{55} & \ding{55} & \ding{55} & 73.42 &92.55 & 40.66 & 95.00 & \textbf{92.92} & 35.86 & 55.29 & \textbf{25.13} & 25.84 & 28.23 \\ 
\method w/o OT & \ding{55} & \ding{51} & \ding{51} & 75.82 & 95.57 & 53.06 & 97.80 & 90.52 & 39.51 & 59.07 & 24.27 & 26.03 & 28.26 \\ 
\method w/ $\mathcal{R}_{\textrm{OT-quality}}$ & \ding{51} & \ding{55} & \ding{51} & 75.97 & 94.84 & 57.80 & \textbf{98.00} & 90.36 & 38.51 & 55.54 & 24.45 & 26.37 & 28.62 \\ 
\method w/ $\mathcal{R}_{\textrm{OT-semantic}}$ & \ding{51} & \ding{51} & \ding{55} & 76.99 & 95.32 & 59.36 & 96.60 & 91.06 & \textbf{44.97} & \textbf{59.93} & 24.04 & 25.81 & 28.26 \\ 
\method & \ding{51} & \ding{51} & \ding{51} &\textbf{77.63} & \textbf{98.13} & \textbf{66.51} & 97.60 & 92.46 & 36.07 & 58.75 & 23.53 & \textbf{26.97} & \textbf{29.10} \\
\bottomrule
\end{tabular}
}
\label{tab:ablation-full}
\end{center}
\end{table*}

\begin{table*}[t]
\centering
\caption{\textbf{Ablation study on Partial OT and spatio-temporal constraints.} 
We report Video-Text Matching (VTM) accuracy on $10{,}000$ video-text pairs from WebVid10M. 
Partial OT with $m=0.9$ achieves the best alignment, while our spatio-temporal constraints 
($\gamma=0.2$, $\eta=0.2$) further boost performance. 
Arrows indicate relative change compared to vanilla cross-attention baseline.}
\label{tab:partial_ot}
\resizebox{0.8\linewidth}{!}{
\begin{tabular}{lcc}
\toprule
\textbf{Method} & \textbf{VTM Acc. (\%)} $\uparrow$& \textbf{Change} \\
\midrule
Vanilla cross-attention & 81.25 & -- \\
Full OT ($m=1.0$) & 86.87 & {\scriptsize \color{ForestGreen}$\uparrow$5.62} \\
 Partial OT ($m=0.5$) & 78.92 & {\scriptsize \color{red}$\downarrow$2.33} \\
 Partial OT ($m=0.9$) & 87.54 & {\scriptsize \color{ForestGreen}$\uparrow$6.29} \\
 Partial OT ($m=0.9$) + spatial only ($\eta=0.2$) & 88.17 & {\scriptsize \color{ForestGreen}$\uparrow$6.92} \\
 Partial OT ($m=0.9$) + temporal only ($\gamma=0.2$) & 87.98 & {\scriptsize \color{ForestGreen}$\uparrow$6.73} \\
 Partial OT ($m=0.9$) + spatio-temporal ($\gamma=\eta=0.1$) & 87.06 & {\scriptsize \color{ForestGreen}$\uparrow$5.81} \\
 Partial OT ($m=0.9$) + spatio-temporal ($\gamma=\eta=0.2$) & \textbf{89.36} & {\scriptsize \color{ForestGreen}$\uparrow$\textbf{8.11}} \\
\bottomrule
\end{tabular}
}
\end{table*}
\section{Optimal Transport Plan Analysis}
\label{sec:pot_analysis}

Table~\ref{tab:partial_ot} analyzes the effect of Partial OT and spatio-temporal constraints on
video-text matching within InternVideo2. We evaluate on $10{,}000$ video-text pairs sampled from
WebVid10M~\citep{bain2021frozen}. 

We observe that using Partial OT with a mass parameter $m=0.9$ achieves the best score of
\textbf{89.36\%}, improving by \textbf{+8.11\%} over vanilla cross-attention. This indicates that not
all text tokens need to be matched to visual patches. For example, uninformative words
(\textit{e.g.}, articles or stopwords) need not be explicitly grounded in the visual domain. Allowing
partial transport filters out such noisy matches while preserving key semantic correspondences.
Conversely, setting $m=0.5$ removes too many tokens, causing essential words to be ignored and
leading to degraded alignment. 

Regarding constraints, our designed cost function with both spatial ($\eta=0.2$) and temporal
($\gamma=0.2$) penalties yields the highest performance, boosting video-text matching by
\textbf{8.11\%} (from 81.25\% to 89.36\%). This demonstrates that integrating spatio-temporal
structure into OT provides sharper and more accurate token-level correspondences, thereby enhancing
fine-grained text-video alignment. Overall, these results confirm the effectiveness of Partial OT
with structured constraints in improving alignment quality.


\section{Motion Guidance in T2V Post-Training}
\label{sec:motion_guide}
For a fair comparison to highlight the impact of rewards, we provide addition ablations (see Table~\ref{tab:motion_guidance}) where we evaluate both methods with and without motion guidance. Without motion guidance, T2V-Turbo-v2~\citep{li2025tvturbov2} performance drops to 83.26 (Quality) and 76.30 (Semantic) on VideoCrafter2~\citep{chen2024videocrafter2}. In comparison, \method-direct still achieves stronger quality (83.73, +0.47) and significantly higher semantic alignment (77.63, +1.33), highlighting the effectiveness of our OT-aligned rewards. Conversely, by adding motion guidance to \method-direct makes it outperform T2V-Turbo-v2 \cite{li2025tvturbov2} across all metrics on VideoCrafter2 \cite{chen2024videocrafter2}.

More notably, on HunyuanVideo~\citep{kong2025hunyuanvideosystematicframeworklarge}, \method-direct \textit{without motion guidance} already \textbf{outperforms T2V-Turbo-v2 with motion} in all metrics (85.05 vs 84.50 Total Score), and this advantage is also reflected in our human preference study (Figure~\ref{fig:preference}). This clearly demonstrates that our reward formulation, rather than the optimization strategy or motion module, is the key driver behind the improvements.

\begin{table*}[!t]
\caption{\small \textbf{Effect of Motion Guidance on post-training VideoCrafter2 and HunyuanVideo}. \method significantly outperforms T2V-Turbo-v2~\citep{li2025tvturbov2} in both with and without motion guidance settings.}
\centering
\setlength\tabcolsep{4pt}

\resizebox{\linewidth}{!}{
\begin{tabular}{@{}l|c@{~~}c@{~~}c|c@{~~}c@{~~}c@{}}
\toprule
\multirow{2}{*}{\textbf{Models}} & \multicolumn{3}{c|}{\textbf{VideoCrafter2}~\citep{chen2024videocrafter2}} & \multicolumn{3}{c}{\textbf{HunyuanVideo}~\citep{kong2025hunyuanvideosystematicframeworklarge}} \\
 & \textbf{Total} & \textbf{Quality} & \textbf{Semantic} & \textbf{Total} & \textbf{Quality} & \textbf{Semantic} \\
\midrule
Vanilla &80.44\xspace\xspace\xspace\xspace\xspace\xspace\xspace\xspace&82.20\xspace\xspace\xspace\xspace\xspace\xspace\xspace\xspace&73.42\xspace\xspace\xspace\xspace\xspace\xspace\xspace\xspace&83.24\xspace\xspace\xspace\xspace\xspace\xspace\xspace\xspace&85.09\xspace\xspace\xspace\xspace\xspace\xspace\xspace\xspace&75.82\xspace\xspace\xspace\xspace\xspace\xspace\xspace\xspace\xspace\\
\midrule
T2V-Turbo-v2 w/o motion~\citep{li2025tvturbov2} 
& 81.87 {\scriptsize \color{ForestGreen}$\uparrow$1.43} 
& 83.26 {\scriptsize \color{ForestGreen}$\uparrow$1.06} 
& 76.30 {\scriptsize \color{ForestGreen}$\uparrow$2.88} 
& 84.25 {\scriptsize \color{ForestGreen}$\uparrow$1.01} 
& 85.93 {\scriptsize \color{ForestGreen}$\uparrow$0.84} 
& 77.52 {\scriptsize \color{ForestGreen}$\uparrow$1.70} \\
\method (Direct Backpropagation) w/o motion & \textbf{82.51} {\scriptsize \color{ForestGreen}$\uparrow$\textbf{2.07}} & \textbf{83.73} {\scriptsize \color{ForestGreen}$\uparrow$\textbf{1.53}} & \textbf{77.63} {\scriptsize \color{ForestGreen}$\uparrow$\textbf{4.21}} & \textbf{85.05} {\scriptsize \color{ForestGreen}$\uparrow$\textbf{1.81}} & \textbf{86.84} {\scriptsize \color{ForestGreen}$\uparrow$\textbf{1.75}} & \textbf{77.89} {\scriptsize \color{ForestGreen}$\uparrow$\textbf{2.07}} \\

\midrule

T2V-Turbo-v2 w/ motion~\citep{li2025tvturbov2} 
& 82.34 {\scriptsize \color{ForestGreen}$\uparrow$1.90} 
& 83.93 {\scriptsize \color{ForestGreen}$\uparrow$1.73} 
& 75.97 {\scriptsize \color{ForestGreen}$\uparrow$2.55} 
& 84.50 {\scriptsize \color{ForestGreen}$\uparrow$1.26} 
& 86.32 {\scriptsize \color{ForestGreen}$\uparrow$1.23} 
& 77.24 {\scriptsize \color{ForestGreen}$\uparrow$1.42} \\

\method (Direct Backpropagation) w/ motion & \textbf{82.79} {\scriptsize \color{ForestGreen}$\uparrow$\textbf{2.35}} & \textbf{84.12} {\scriptsize \color{ForestGreen}$\uparrow$\textbf{1.92}} & \textbf{77.45} {\scriptsize \color{ForestGreen}$\uparrow$\textbf{4.03}} & \textbf{85.24} {\scriptsize \color{ForestGreen}$\uparrow$\textbf{2.00}} & \textbf{87.07} {\scriptsize \color{ForestGreen}$\uparrow$\textbf{1.98}} & \textbf{77.94} {\scriptsize \color{ForestGreen}$\uparrow$\textbf{2.12}} \\
\bottomrule
\end{tabular}
}
\label{tab:motion_guidance}
\end{table*}

\section{\method with ViCLIP Evaluator}
\label{sec:viclip_evaluator}
\begin{table*}[!t]
\caption{\small \textbf{Comparison of Evaluators for \method on VBench}. Both InternVideo2 and ViCLIP as evaluators yield strong improvements over the Vanilla baseline, confirming \method' robustness across feature extractors.}
\centering
\setlength\tabcolsep{4pt}

\resizebox{\linewidth}{!}{
\begin{tabular}{@{}l|c@{~~}c@{~~}c|c@{~~}c@{~~}c@{}}
\toprule
\multirow{2}{*}{\textbf{Models}} & \multicolumn{3}{c|}{\textbf{VideoCrafter2}~\citep{chen2024videocrafter2}} & \multicolumn{3}{c}{\textbf{HunyuanVideo}~\citep{kong2025hunyuanvideosystematicframeworklarge}} \\
 & \textbf{Total} & \textbf{Quality} & \textbf{Semantic} & \textbf{Total} & \textbf{Quality} & \textbf{Semantic} \\
\midrule
Vanilla &80.44\xspace\xspace\xspace\xspace\xspace\xspace\xspace\xspace&82.20\xspace\xspace\xspace\xspace\xspace\xspace\xspace\xspace&73.42\xspace\xspace\xspace\xspace\xspace\xspace\xspace\xspace&83.24\xspace\xspace\xspace\xspace\xspace\xspace\xspace\xspace&85.09\xspace\xspace\xspace\xspace\xspace\xspace\xspace\xspace&75.82\xspace\xspace\xspace\xspace\xspace\xspace\xspace\xspace\xspace\\
\method w/ InternVideo2~\citep{wang2024internvideo2} & 82.75 {\scriptsize \color{ForestGreen}$\uparrow$2.31} & 84.05 {\scriptsize \color{ForestGreen}$\uparrow$1.85} & 77.54 {\scriptsize \color{ForestGreen}$\uparrow$4.12} & 85.45 {\scriptsize \color{ForestGreen}$\uparrow$2.21} & 86.73 {\scriptsize \color{ForestGreen}$\uparrow$1.64} & 80.33 {\scriptsize \color{ForestGreen}$\uparrow$4.51} \\
\method w/ ViCLIP~\citep{wang2023viclip} & 82.84 {\scriptsize \color{ForestGreen}$\uparrow$2.40} & 84.18 {\scriptsize \color{ForestGreen}$\uparrow$1.98} & 77.49 {\scriptsize \color{ForestGreen}$\uparrow$4.07} & 85.33 {\scriptsize \color{ForestGreen}$\uparrow$2.09} & 86.77 {\scriptsize \color{ForestGreen}$\uparrow$1.68} & 79.58 {\scriptsize \color{ForestGreen}$\uparrow$3.76} \\
\bottomrule
\end{tabular}
}
\label{tab:evaluator-diversity}
\end{table*}
To test whether \method improvements generalize beyond InternVideo2~\citep{wang2024internvideo2}, we conducted a controlled experiment where both our rewards (Distributional OT and Semantic OT) are based on ViCLIP~\citep{wang2023viclip}, a CLIP-based video–text encoder that differs from InternVideo2 in training corpus and representation space. The results in Table~\ref{tab:evaluator-diversity} confirm that \method remains effective even when rewards and evaluation use ViCLIP \cite{wang2023viclip}, achieving comparable performance across both benchmarks and further mitigating concerns of reward overfitting to a specific model family. These consistent results show the generality of our OT-aligned reward formulation and demonstrate that the observed improvements are not confined to InternVideo2 features.

\section{Scope of Generalization}
\label{sec:ood_prompts}
To assess the robustness of \method beyond the WebVid10M~\citep{bain2021frozen} and VidGen-1M~\citep{tan2024vidgen1mlargescaledatasettexttovideo} domain, we curated 100 diverse out-of-distribution (OOD) prompts. These prompts cover challenging and underrepresented scenarios including robotics actions, embodied tasks, procedural instructions, abstract concepts, and rare object-event compositions. Example prompts include:
\begin{itemize}
    \item \textit{A robot arm with a red gripper picks up a blue cube and sorts it into a green bin on a moving conveyor belt in a bright factory hall.}
    \item \textit{A tiny mechanical mouse navigates through a labyrinth of gears inside a giant clock, camera close-up on its delicate paws.}
\end{itemize}
To test alignment under these harder OOD conditions, we compare the cosine similarity between text prompt embeddings and generated video embeddings using the ViCLIP~\citep{wang2023viclip} encoder. Results in Table~\ref{table: ood_prompts} confirm that \method exhibits stronger alignment even under distribution shifts and when evaluated using an independent video-text encoder ViCLIP~\citep{wang2023viclip}, supporting the generality and robustness of our OT-aligned rewards.

\begin{table}[!t]
\centering
\caption{\small \textbf{OOD Alignment Performance using ViCLIP~\citep{wang2023viclip}.} \method achieves the highest cosine similarity, confirming robustness under distribution shifts.}
\resizebox{\linewidth}{!}{
\begin{tabular}{l|c}
\toprule
\textbf{Method} & \textbf{Cosine Similarity} $\uparrow$ \\
\midrule
HunyuanVideo~\citep{kong2025hunyuanvideosystematicframeworklarge} & 0.4128 \\
T2V-Turbo-v2~\citep{li2024t2vturbo} & 0.4390 \\
VideoReward-DPO~\citep{liu2025improving} & 0.4285 \\
\method & \textbf{0.4517} \\
\bottomrule
\end{tabular}
}
\label{table: ood_prompts}
\end{table}

\section{Hyperparameters Sensitivity Analysis}
\label{sec:hyper_sensitivity}
To further assess robustness, we conduct a stability analysis by sweeping $(\gamma, \eta)$ over the range $[0.0, 0.5]$ in $0.1$ increments. For each setting, we measure the Video-Text Matching (VTM) accuracy on $10{,}000$ WebVid10M~\citep{bain2021frozen} video-text pairs using our OT-aligned Semantic Reward with InternVideo2~\citep{wang2024internvideo2}.

\begin{figure}[t]
  \centering
  \includegraphics[width=\linewidth]{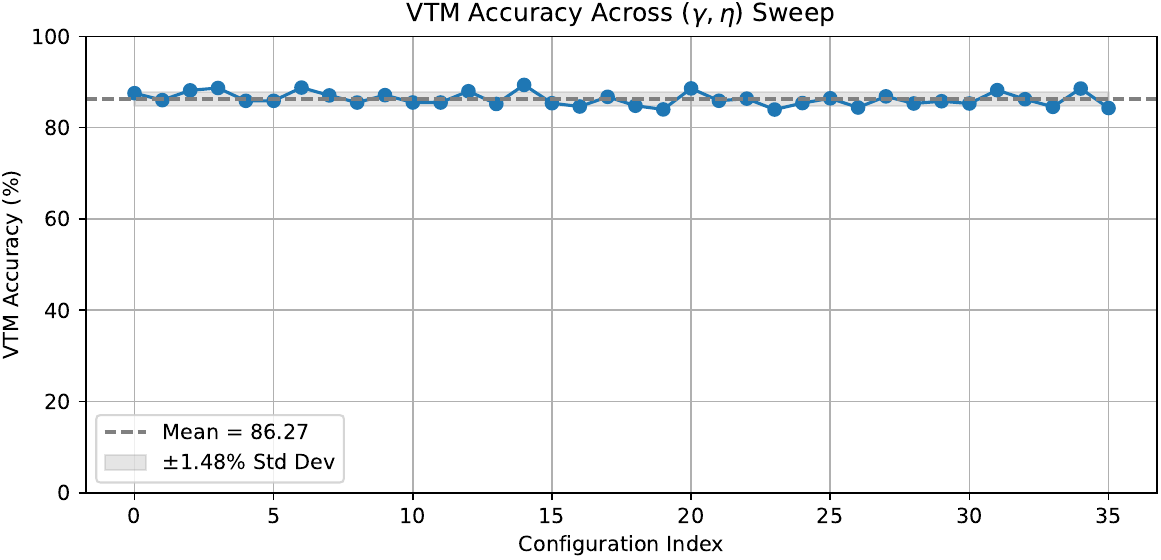} 
  \caption{\small VTM accuracy under $(\gamma,\eta)$ sweep. Despite configuration variations, the accuracy remains stable ($\pm$1.48\% std).}
  \label{fig:sweep_std}
\end{figure}

We observe that the VTM accuracy varies smoothly across this range, with a standard deviation of only $1.48\%$, indicating that our method is robust to these weights. This confirms the stability of our Partial OT formulation with respect to its structured penalty terms. As shown in Figure~\ref{fig:sweep_std}, the variation is minimal and has a negligible impact on performance.

\section{Failure Example of Discrete Partial OT}
\label{sec:failure_mode}
\begin{figure}[t!]
\centering
\includegraphics[width=\linewidth]{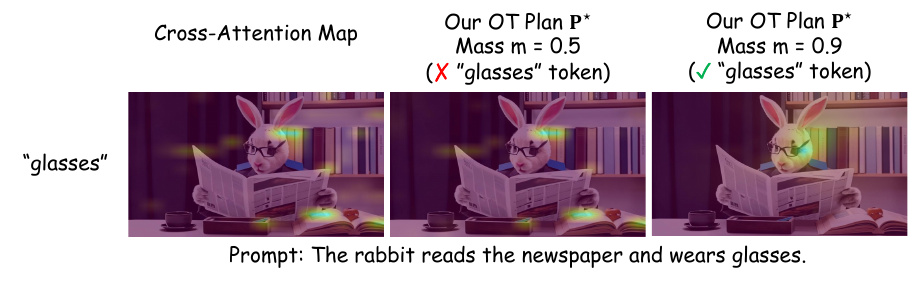}
\caption{\textbf{Effect of OT mass parameter.} Low OT mass ($m=0.5$) fails to retain the \textit{“glasses”} token, while higher mass ($m=0.9$) improves alignment.}
\label{fig:failure-case}
\end{figure}

We provide a qualitative example in Figure~\ref{fig:failure-case} that illustrates both the sensitivity and limitations of OT alignment. \textbf{Left:} The attention map from vanilla cross-attention fails to ground the token \textit{“glasses”}. \textbf{Middle:} Our OT plan $\mathbf{P}^\star$ with partial mass $m = 0.5$ suppresses noise but also removes the valid token \textit{“glasses”}, reproducing misalignment. \textbf{Right:} With mass $m = 0.9$, the OT plan retains the \textit{“glasses”} token and aligns it better--but still imperfectly, activating only the right lens.

This reveals a current limitation of the setup: while the OT module improves semantic alignment through structured grounding, overall performance depends on the representational precision of the underlying video–text model, InternVideo2~\citep{wang2024internvideo2}, which was not optimized for fine-grained localization tasks such as segmentation. As a result, even a well-structured transport plan may not fully resolve detailed grounding errors when the base features do not capture sufficient spatial detail. Our approach remains agnostic to the underlying base model, and continued progress in open-source VLMs is likely to further enhance fine-grained grounding performance.

\section{Human Preference Details}
\label{sec:human_details}
To assess inter-rater reliability, we computed Fleiss’ Kappa over the collected human preference judgments. Across the three evaluation axes (visual quality, motion quality, and semantic alignment), we obtained an average Fleiss’ Kappa score of 0.72, indicating substantial agreement among raters. All raters were blinded to method identity to prevent bias.


We analyzed category-level bias by classifying 400 prompts into 290 motion-heavy (e.g., running, jumping) and 110 static (e.g., portraits, scenic shots). The Pearson correlation between category type and preference for our method is 0.038, indicating no significant bias--our method performs consistently across both motion-heavy and static prompts.

\begin{figure*}[t!]
\centering
\includegraphics[width=\linewidth]{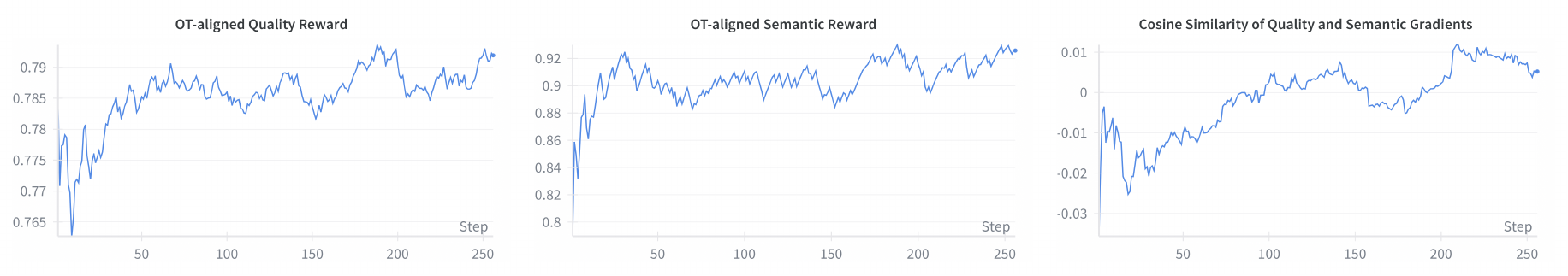}
\caption{Reward and gradient trends in post-training.}
\label{fig:reward_trend}
\end{figure*}

\section{Reward Fusion}
\label{sec:reward_fusion}
In our default direct backpropagation setup, we equally weight the consistency loss and reward signals. Moreover, we explore an adaptive weighting strategy using the Group-Relative Reward formulation from GRPO~\citep{Guo2025DeepSeekR1, liu2025flowgrpo}, normalizing reward values across generations for the same prompt (subtracting the mean and dividing by standard deviation). We apply this normalization in the direct backpropagation setting without using RL. Results on HunyuanVideo~\citep{kong2025hunyuanvideosystematicframeworklarge} (Table~\ref{tab:adaptive-weighting}) indicate that adaptive weighting via Group-Relative Reward offers a consistent improvement over equal weighting, particularly in semantic alignment and overall score, confirming its potential as a robust reward fusion strategy.

\input{tables/adaptive_reward_fusion}

\section{Dual Rewards Interaction}
\label{sec:reward_interaction}

To assess the interaction between the OT-aligned quality and semantic rewards during training, we plot their reward trajectories and the cosine similarity of their gradients over time in Figure~\ref{fig:reward_trend}. We observe that both rewards improve steadily across training steps. At step 256 (final), the semantic reward reaches a high value of $0.9268$, and the quality reward converges to $0.7814$, suggesting the model successfully learns to optimize for both objectives. The cosine similarity of the gradients between the two rewards remains near zero throughout training (final value $0.0074$), with small fluctuations in both directions. This implies that the two objectives provide largely orthogonal supervision signals and do not interfere with each other. There is no sign of reward conflict or mode collapse. Together, these findings indicate that our dual reward formulation, one distributional and one token-level, are both stable. It effectively supports the joint optimization of visual quality and semantic alignment in text-to-video generation. This stability enables consistent improvements across both dimensions without sacrificing one for the other.

\section{Reward Hacking and Mitigation Strategy}
\label{sec:reward_hacking}
We found that the Consistency Distillation (CD) \cite{song2023consistency, wang2023videolcm, lu2025simplifying} loss helps mitigate reward hacking and stabilize training. Intuitively, this loss anchors the student model (updated via reward supervision) to the teacher model’s original distribution, serving both as a regularizer and as an efficient training mechanism by enabling fewer denoising steps. We evaluate the impact of CD loss in Table~\ref{tab:cd-loss}. Without CD loss, the semantic reward is optimized more aggressively, leading to a noticeable drop in visual quality (from 86.84 to 86.51) and total score (from 85.05 to 84.80), confirming that the model may over-optimize the reward signal at the expense of generation fidelity. By retaining CD loss, we balance the original model performance and reward-driven improvements.

\begin{table}[!t]
\caption{\small \textbf{Impact of CD Loss on Reward Stability and Quality.} CD loss serves as a regularizer that prevents overfitting to reward signals and stabilizes training.}
\centering
\setlength\tabcolsep{4pt}
\resizebox{\linewidth}{!}{
\begin{tabular}{@{}l|c@{~~}c@{~~}c@{}}
\toprule
\textbf{Method} & \textbf{Total Score} & \textbf{Quality} & \textbf{Semantic} \\
\midrule
HunyuanVideo~\citep{kong2025hunyuanvideosystematicframeworklarge} & 83.24\xspace\xspace\xspace\xspace\xspace\xspace\xspace\xspace & 85.09\xspace\xspace\xspace\xspace\xspace\xspace\xspace\xspace & 75.82\xspace\xspace\xspace\xspace\xspace\xspace\xspace\xspace \\\method w/o CD Loss & 84.80 {\scriptsize \color{ForestGreen}$\uparrow$1.56} & 86.51 {\scriptsize \color{ForestGreen}$\uparrow$1.42} & \textbf{77.95 {\scriptsize \color{ForestGreen}$\uparrow$2.13}} \\
\method w/ CD Loss & \textbf{85.05 {\scriptsize \color{ForestGreen}$\uparrow$1.81}} & \textbf{86.84 {\scriptsize \color{ForestGreen}$\uparrow$1.75}} & 77.89 {\scriptsize \color{ForestGreen}$\uparrow$2.07} \\
\bottomrule
\end{tabular}
}
\label{tab:cd-loss}
\end{table}

\section{Training Cost and Efficiency Analysis}
\label{sec:training_cost}
To train the OT map \cite{korotin2023neural}, we use video-text pairs with 8-frame clips, extracted using frozen InternVideo2 \cite{wang2024internvideo2}, and train on a single A100 GPU for one day, equivalent to 24 A100 GPU-hours. In comparison, annotation-based pipelines such as VideoReward-DPO \cite{liu2025improving} require 72 A800 GPU-hours to train their reward models.

The total wall-clock cost of \method post-training is 29.78 GPU-hours on 8$\times$A100s, slightly higher than the 26.52 GPU-hours required by T2V-Turbo-v2 \cite{li2025tvturbov2} (training time only, without evaluation). While \method incurs a marginal increase in training time, this is a one-time cost and remains negligible compared to the massive pre-training costs of T2V models-665,000 GPU-hours for Seaweed-7B \cite{seawead2025seaweed7bcosteffectivetrainingvideo} and even more for HunyuanVideo-13B \cite{kong2025hunyuanvideosystematicframeworklarge}. Thus, \method provides a lightweight and effective reward alignment mechanism with minimal computational overhead.

Finally, inference incurs no additional cost from the reward models. In fact, for GRPO-tuned models, we reduce the denoising steps from 50 to 16 through consistency distillation, resulting in approximately 3$\times$ faster inference.

%% file: tables/adaptive_reward_fusion.tex
\begin{table}[!t]
\caption{\small \textbf{Adaptive Reward Fusion}. We compare equal and adaptive weighting of reward signals in direct backprop. Adaptive weighting using Group Relative Reward improves performance.}
\centering
\resizebox{\linewidth}{!}{
\begin{tabular}{@{}l|c@{~~}c@{~~}c@{}}
\toprule
\textbf{Method} & \textbf{Total} & \textbf{Quality} & \textbf{Semantic} \\
\midrule
HunyuanVideo & 83.24\xspace\xspace\xspace\xspace\xspace\xspace\xspace\xspace & 85.09\xspace\xspace\xspace\xspace\xspace\xspace\xspace\xspace & 75.82\xspace\xspace\xspace\xspace\xspace\xspace\xspace\xspace \\
\method-direct (equal) & 85.05 {\scriptsize \color{ForestGreen}$\uparrow$1.81} & 86.84 {\scriptsize \color{ForestGreen}$\uparrow$1.75} & 77.89 {\scriptsize \color{ForestGreen}$\uparrow$2.07} \\
\method-direct (adaptive) & \textbf{85.16} {\scriptsize \color{ForestGreen}$\uparrow$\textbf{1.92}} & \textbf{86.92} {\scriptsize \color{ForestGreen}$\uparrow$\textbf{1.83}} & \textbf{78.11} {\scriptsize \color{ForestGreen}$\uparrow$\textbf{2.29}} \\
\bottomrule
\end{tabular}
}
\label{tab:adaptive-weighting}
\end{table}